\documentclass[10pt,twocolumn,letterpaper]{article}

\usepackage{cvpr}
\usepackage{times}
\usepackage{epsfig}
\usepackage{graphicx}
\usepackage{amsmath}
\usepackage{amssymb}

\usepackage{tabularx} 
\newcolumntype{Y}{>{\centering\arraybackslash}X}

\usepackage{array,booktabs,arydshln,xcolor} 
\usepackage{amsthm,bm}
\usepackage{epstopdf}
\usepackage[flushleft]{threeparttable}
\usepackage{amsthm}
\usepackage{multirow}

\usepackage{float}
\usepackage{wrapfig}
\usepackage{subcaption}
\usepackage{capt-of}
\usepackage{booktabs}
\usepackage{varwidth}
\usepackage{wrapfig}
\usepackage{makecell}

\usepackage{algpseudocode,algorithm,algorithmicx}

\algrenewcommand\algorithmicrequire{\textbf{Precondition:}}
\algrenewcommand\algorithmicensure{\textbf{Postcondition:}}

\makeatletter
\def\BState{\State\hskip-\ALG@thistlm}
\makeatother

\usepackage{bm}

\usepackage[pagebackref=true,breaklinks=true,letterpaper=true,colorlinks,bookmarks=false]{hyperref}

\cvprfinalcopy 

\def\cvprPaperID{2823} 

\begin{document}

\title{Photographic Text-to-Image Synthesis \\ with a Hierarchically-nested Adversarial Network}

\author{Zizhao Zhang$^*$, ~Yuanpu Xie$^*$, ~Lin Yang \\
		University of Florida \\
		{\tt\small $^*$equal contribution} \\
}

\maketitle

\begin{abstract}
This paper presents a novel method to deal with the challenging task of generating photographic images conditioned on semantic image descriptions.
Our method introduces accompanying hierarchical-nested adversarial objectives inside the network hierarchies, which regularize mid-level representations and assist generator training to capture the complex image statistics. We present an extensile single-stream generator architecture to better adapt the jointed discriminators and push generated images up to high resolutions. We adopt a multi-purpose adversarial loss to encourage more effective image and text information usage in order to improve the semantic consistency and image fidelity simultaneously. Furthermore, we introduce a new visual-semantic similarity measure to evaluate the semantic consistency of generated images. With extensive experimental validation on three public datasets, our method significantly improves previous state of the arts on all datasets over different evaluation metrics.

\end{abstract}

\section{Introduction}
Photographic text-to-image synthesis is a significant problem in generative model research \cite{reed2016generative}, which aims to learn a mapping from a semantic text space to a complex RGB image space. This task requires the generated images to be not only realistic but also \textit{semantically consistent}, i.e., the generated images should preserve specific object sketches and semantic details described in text.




\begin{figure}[t]
    \centering
    \begin{subfigure}[t]{0.5\textwidth}
        \includegraphics[width=0.95\textwidth]{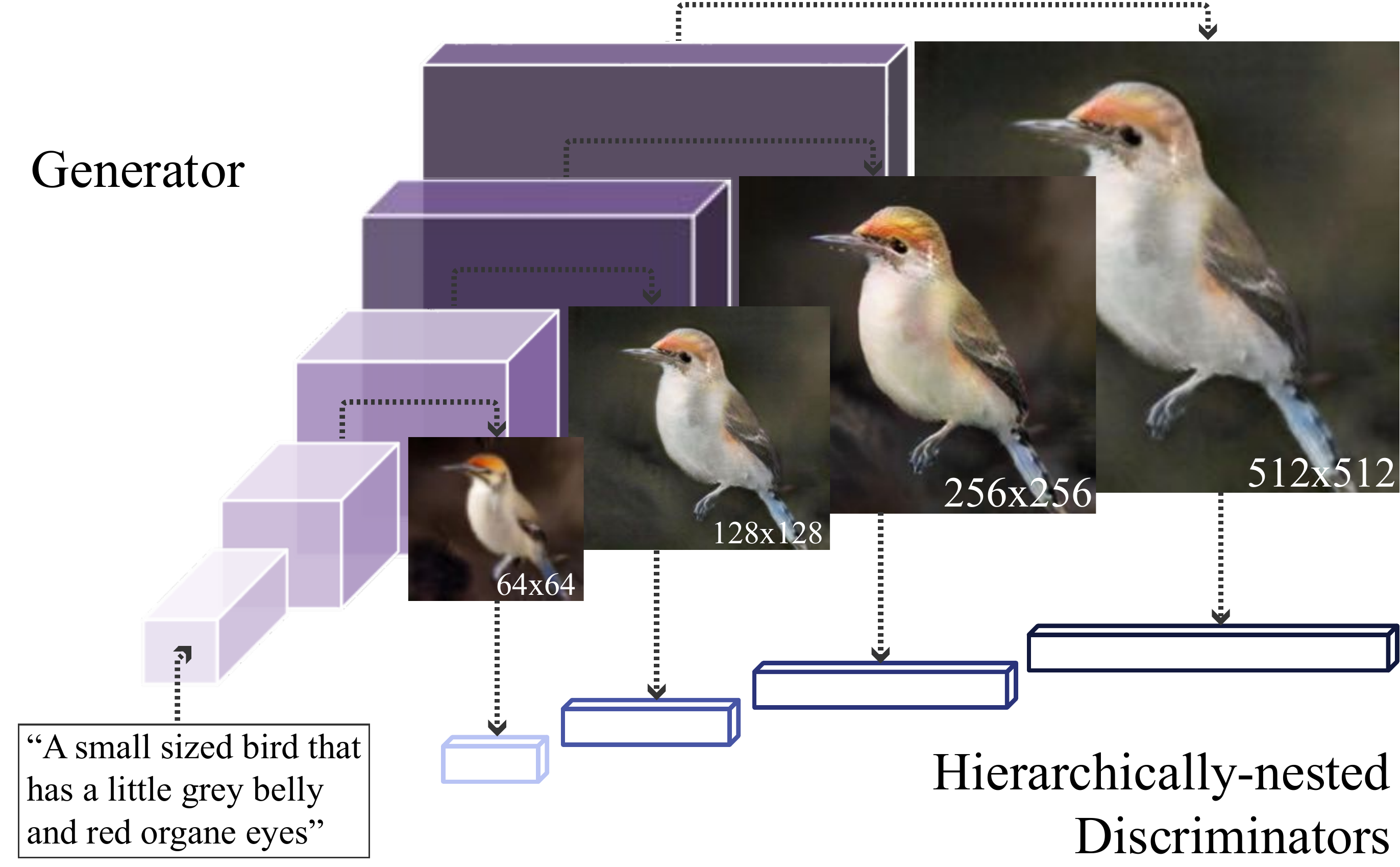}
    \end{subfigure} \\
\centering
    \begin{subfigure}[t]{0.5\textwidth}
        \includegraphics[width=0.95\textwidth,height=0.55\textwidth]{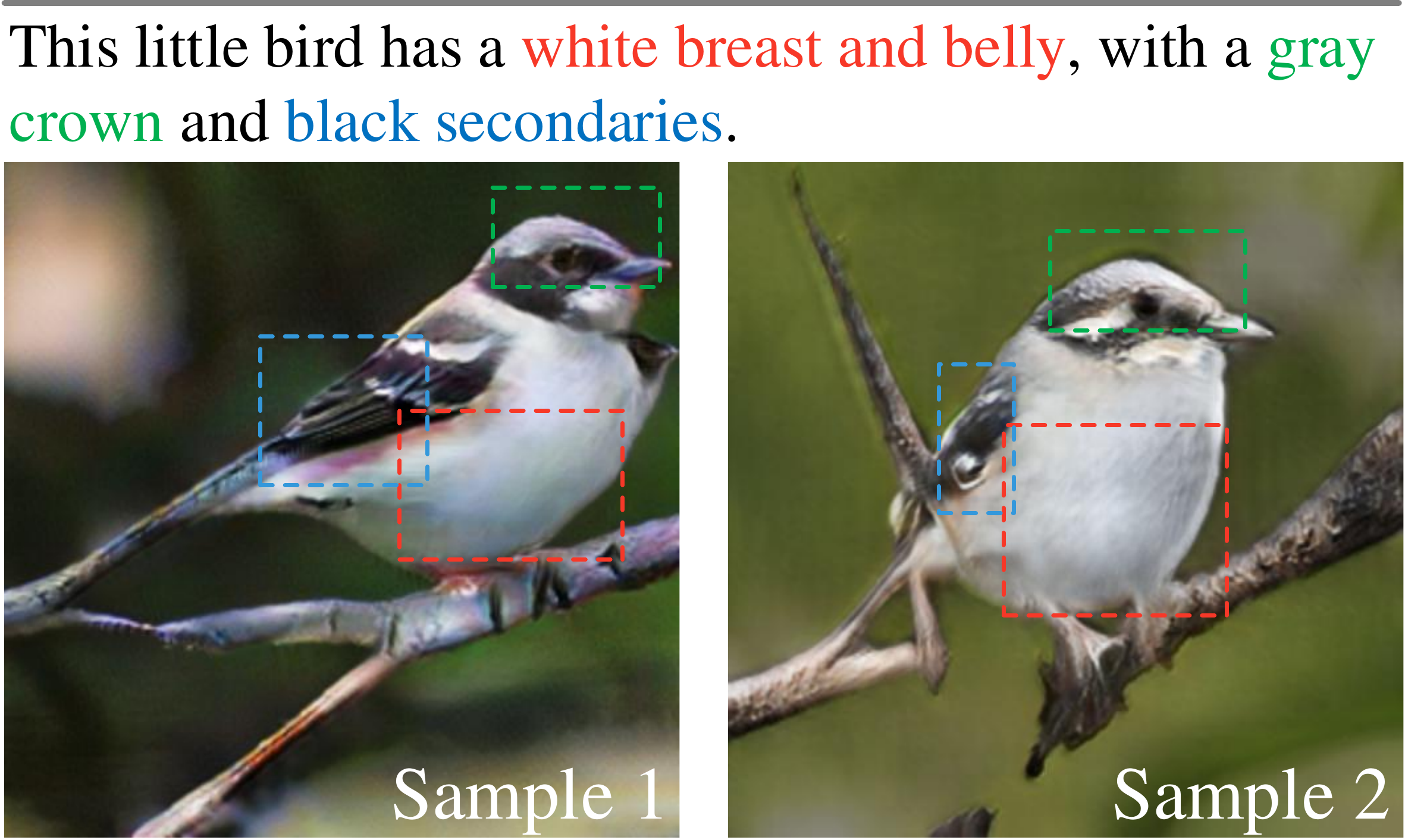}
    \end{subfigure}
    \vspace{-.6cm}
    \caption{Top: Overview of our hierarchically-nested adversarial network, which produces side output images with growing resolutions. Each side output is associated with a discriminator. Bottom: Two test sample results where fine-grained details are highlighted. \label{fig:intro}}     
    \vspace{-.5cm}
\end{figure}


Recently, Generative adversarial networks (GANs) have become the main solution to this task. 
Reed \etal \cite{reed2016generative} address this task through a GAN based framework. But this method only generates $64^2$ images and can barely generate vivid object details.
Based on this method, StackGAN \cite{han2017stackgan} proposes to stack another low-to-high resolution GAN to generate $256^2$ images. But this method requires training two separate GANs. Later on, \cite{dong2017semantic} proposes to bypass the difficulty of learning mappings from text to RGB images and treat it as a pixel-to-pixel translation problem \cite{isola2016image}. It works by re-rendering an arbitrary-style $128^2$ training image conditioned on a targeting description. 
However, its high-resolution synthesis capability is unclear. 
At present, training a generative model to map from a low-dimensional text space to a high-resolution image space in a fully end-to-end manner still remains unsolved. 

This paper pays attention to two major difficulties for text-to-image synthesis with GANs. The first is balancing the convergence between generators and discriminators \cite{goodfellow2014generative,improvedGAN}, which is a common problem in GANs. The second is stably modeling the huge pixel space in high-resolution images and guaranteeing semantic consistency \cite{han2017stackgan}. 
An effective strategy to regularize generators is critical to stabilize the training and help capture the complex image statistics \cite{huang2016stacked}. 

In this paper, we propose a novel end-to-end method that can directly model high-resolution image statistics and generate photographic images (see Figure \ref{fig:intro} bottom). The contributions are described as follows.

Our generator resembles a simple vanilla GAN, without requiring multi-stage training and multiple internal text conditioning like \cite{han2017stackgan} or additional class label supervision like \cite{dash2017tac}. To tackle the problem of the big leap from the text space to the image space, our insight is to leverage and regularize hierarchical representations with additional `deep' adversarial constraints (see Figure \ref{fig:intro} top). 
We introduce accompanying hierarchically-nested discriminators at multi-scale intermediate layers to play adversarial games and jointly encourage the generator to approach the real training data distribution. 
We also propose a new convolutional neural network (CNN) design for the generator to support accompanying discriminators more effectively.
To guarantee the image diversity and semantic consistency, we enforce discriminators at multiple side outputs of the generator to simultaneously differentiate real-and-fake image-text pairs as well as real-and-fake local image patches. 


We validate our proposed method on three datasets, CUB birds \cite{welinder2010caltech}, Oxford-102 flowers \cite{Nilsback08}, and large-scale MSCOCO \cite{lin2014microsoft}. In complement of existing evaluation metrics (e.g. Inception score \cite{improvedGAN}) for generative models, we also introduce a new visual-semantic similarity metric to evaluate the alignment between generated images and conditioned text. It alleviates the issue of the expensive human evaluation. Extensive experimental results and analysis demonstrate the effectiveness of our method and significantly improved performance compared against previous state of the arts on all three evaluation metrics. 
All source code will be released.

\begin{figure*}[t]
    \centering
    \includegraphics[width=0.95\textwidth]{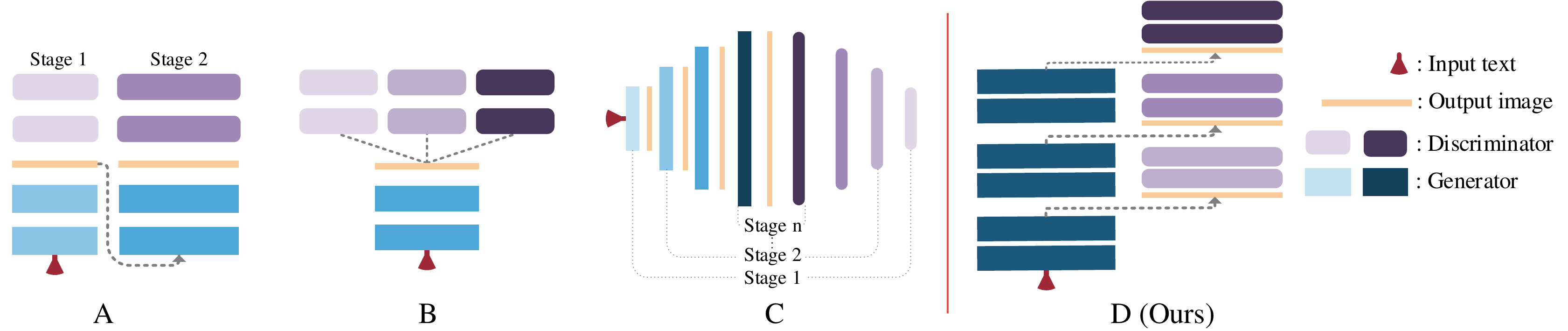}
    \vspace{-.2cm}
    \caption{Overviews of some typical GAN frameworks. \textbf{A} uses multi-stage GANs \cite{han2017stackgan,denton2015deep}. \textbf{B} uses multiple discriminators with one generator \cite{durugkar2016generative,tu_etal_nips17_d2gan}. \textbf{C} progressively trains symmetric discriminators and generators \cite{Karras2017progressive,huang2016stacked}. \textbf{A} and \textbf{C} can be viewed as decomposing high-resolution tasks to multi-stage low-to-high resolution tasks.  \textbf{D} is our proposed framework that uses a single-stream generator with hierarchically-nested discriminators trained end-to-end.} \label{fig:archs-review} \vspace{-.4cm}
\end{figure*}

\section{Related Work}

Deep generative models have attracted wide interests recently, including GANs \cite{goodfellow2014generative,radford2015unsupervised}, Variational Auto-encoders (VAE) \cite{kingma2013auto}, etc \cite{oord2016pixel}. 
There are substantial existing methods investigating the better usage of GANs for different applications, such as image synthesis \cite{radford2015unsupervised, shrivastava2016learning}, (unpaired) pixel-to-pixel translation \cite{isola2016image,zhu2017unpaired}, medical applications \cite{costa2017towards,zhang2018cardiac}, etc \cite{ledig2016photo,zhang2017multistyle,huang2016stacked,zhang2017image}.

Text-to-image synthesis is an interesting application of GANs. Reed \etal \cite{reed2016generative} is the first to introduce a method that can generate $64^2$ resolution images. This method also presents a new strategy for image-text matching aware adversarial training. Reed \etal \cite{reed2016learning} propose a generative
adversarial what-where network (GAWWN) to enable location and content instructions in text-to-image synthesis. Zhang \etal \cite{han2017stackgan} propose a two-stage training strategy that is able to generate $256^2$ compelling images. Recently, Dong \etal \cite{dong2017semantic} propose to learn a joint embedding of images and text so as to re-render a prototype image conditioned on a targeting description. Cha \etal \cite{char2017perceptual} explore the usage of the perceptional loss \cite{johnson2016perceptual} with a CNN pretrained on ImageNet  and Dash \etal \cite{dash2017tac} make use of auxiliary classifiers (similar with \cite{odena2016conditional}) to assist GAN training for text-to-image synthesis. Xu \etal \cite{xu2017attngan} shows an attention-driven method to improve fine-grained details. 
    
Learning a continuous mapping from a low-dimensional manifold to a complex real image distribution is a long-standing problem. Although GANs have made significant progress, there are still many unsolved difficulties, e.g., training instability and high-resolution generation. Wide methods have been proposed to address the training instability, such as various training techniques \cite{salimans2016improved,berthelot2017began,shrivastava2016learning,odena2016conditional}, regularization using extra knowledge (e.g. image labels, ImageNet CNNs) \cite{dosovitskiy2016generating,ledig2016photo,dash2017tac}, or different generator and discriminator combinations  \cite{metz2016unrolled,durugkar2016generative,huang2016stacked}. \textit{While our method shows a new way to unite generators and discriminators and does not require any extra knowledge apart from training paired text and images.} In addition, it is easy to see the training difficulty increases significantly as the targeting image resolution increases.


To synthesize high-resolution images, cascade networks are effective to decompose originally difficult tasks to multiple subtasks (Figure \ref{fig:archs-review} A).
Denton \etal \cite{denton2015deep} train a cascade of GANs in a Laplacian pyramid framework (LAPGAN) and use each to synthesize and refine image details and push up the output resolution stage-by-stage. StackGAN also shares similar ideas with LAPGAN. Inspired by this strategy, Chen \etal \cite{chen2017photographic} present a cascaded refinement network to synthesize high-resolution scenes from semantic maps. 
Recently, Karras \etal \cite{Karras2017progressive} propose a progressive training of GANs. The idea is to add symmetric generator and discriminator layers gradually for high-resolution image generation (Figure \ref{fig:archs-review} C). \textit{Compared with these strategies that train low-to-high resolution GANs stage-by-stage or progressively, our method has the advantage of leveraging mid-level representations to encourage the integration of multiple subtasks, which makes end-to-end high-resolution image synthesis in a single vanilla-like GAN possible.}


Leveraging hierarchical representations of CNNs is an effective way to enhance implicit multi-scaling and ensembling for tasks such as image recognition \cite{lee2015deeply,Zhang_2017_CVPR} and pixel or object classification \cite{xie2015holistically,long2015fully,zhao2017pyramid}. Particularly, using deep supervision \cite{lee2015deeply} at intermediate convolutional layers provides short error paths and increases the discriminativeness of feature representations. 
\textit{Our hierarchically-nested adversarial objective is inspired by the family of deeply-supervised CNNs.}

\section{Method}

\subsection{Adversarial objective basics}
In brief, a GAN \cite{goodfellow2014generative} consists of a generator $G$ and a discriminator $D$, which are alternatively trained to compete with each other. $D$ is optimized to distinguish synthesized images from real images, meanwhile, $G$ is trained to fool $D$ by synthesizing fake images. Concretely, the optimal $G$ and $D$ can be obtained by playing the following two-player min-max game,
\begin{equation}
\label{equ:GAN}
\begin{split}
G^*, D^* = \arg~\underset{G}{\min}\ \underset{D}{\max}~ \mathcal{V}(D, G, Y, \bm z),
\end{split}
\end{equation}
where $Y$ and $\bm z \sim \mathcal{N}(0,1)$ denote training images and random noises, respectively. 
$\mathcal{V}$ is the overall GAN objective, which usually takes the form $\mathbb{E}_{Y\sim p_{data}}\big[\log D(Y)\big] + \mathbb{E}_{\bm z\sim p_{z}}\big[\log (1-D(G(\bm z)))\big]$ (the cross-entropy loss) or other variations \cite{lsgan,berthelot2017began}.

\subsection{Hierarchical-nested adversarial objectives}
Numerous GAN methods have demonstrated ways to unite generators and discriminators for image synthesis. Figure \ref{fig:archs-review} and Section 2 discuss some typical frameworks.
Our method actually explores a new dimension of playing this adversarial game along the depth of a generator (Figure \ref{fig:archs-review} D), which integrates additional hierarchical-nested discriminators at intermediate layers of the generator. 
The proposed objectives act as regularizers to the hidden space of $\mathcal{G}$, which also offer a short path for error signal flows and help reduce the training instability.

The proposed $\mathcal{G}$ is a CNN (defined in Section 3.4), which produces multiple side outputs:
\begin{equation}
\label{side}
X_1,..., X_s = \mathcal{G}(\bm t, \bm z), 
\end{equation}
where  $\bm t\sim p_{data}$ denotes a sentence embedding (generated by a pre-trained char-RNN text encoder \cite{reed2016generative}). $\{X_1,...,X_{s-1}\}$ are images with gradually growing resolutions and $X_s$ is the final output with the highest resolution.


For each side output $X_i$ from the generator, a distinct discriminator $D_i$ is used to compete with it. Therefore, our full min-max objective is defined as 
\begin{equation}
\label{equ:optim}
\begin{split}
  \mathcal{G}^*, \mathcal{D}^*&  =  \arg~\underset{\mathcal{G}}{\min}\ \underset{\mathcal{D}}{\max}~ \mathcal{V}(\mathcal{G},\mathcal{D}, \mathcal{Y}, \bm t, \bm z), \\
\end{split}
\end{equation}
where $\mathcal{D} = \{D_1, ..., D_s\} $ and $\; \mathcal{Y} = \{Y_1, ..., Y_s\}$ denotes training images at multi-scales, $\{1,...,s\}$.
Compared with Eq. (\ref{equ:GAN}), our generator competes with multiple discriminators $\{D_i\}$ at different hierarchies (Figure \ref{fig:archs-review} D), which jointly learn discriminative features in different contextual scales.

In principle, the lower-resolution side output is used to learn semantic consistent image structures (e.g. object sketches, colors, and background), and the subsequent higher-resolution side outputs are used to render fine-grained details. Since our method is trained in an end-to-end fashion, the lower-resolution outputs can also fully utilize top-down knowledge from discriminators at higher resolutions. As a result, we can observe consistent image structures, color and styles in the outputs of both low and high resolution images. The experiment demonstrates this advantage compared with StackGAN. 


\subsection{Multi-purpose adversarial losses}
Our generator produces resolution-growing side outputs composing an image pyramid. 
We leverage this hierarchy property and allow adversarial losses to capture hierarchical image statistics, with a goal to guarantee both semantic consistency and image fidelity. 

In order to guarantee semantic consistency, we adopt the matching-aware pair loss proposed by \cite{reed2016generative}. The discriminator is designed to take image-text pairs as inputs and be trained to identify two types of errors: a real image with mismatched text and a fake image with conditioned text.

The pair loss is designed to guarantee the global semantic consistency. However, there is no explicit loss to guide the discriminator to differentiate real images from fake images. And combining both tasks (generating realistic images and matching image styles with text) into one network output complicates the already challenging learning tasks. Moreover, as the image resolution goes higher, it might be challenging for a global pair-loss discriminator to capture the local fine-grained details (results are validated in experiments).
In addition, as pointed in \cite{shrivastava2016learning}, a single global discriminator may over-emphasize certain biased local features and lead to artifacts. 

\begin{figure}[t]
    \centering
    \includegraphics[width=0.49\textwidth]{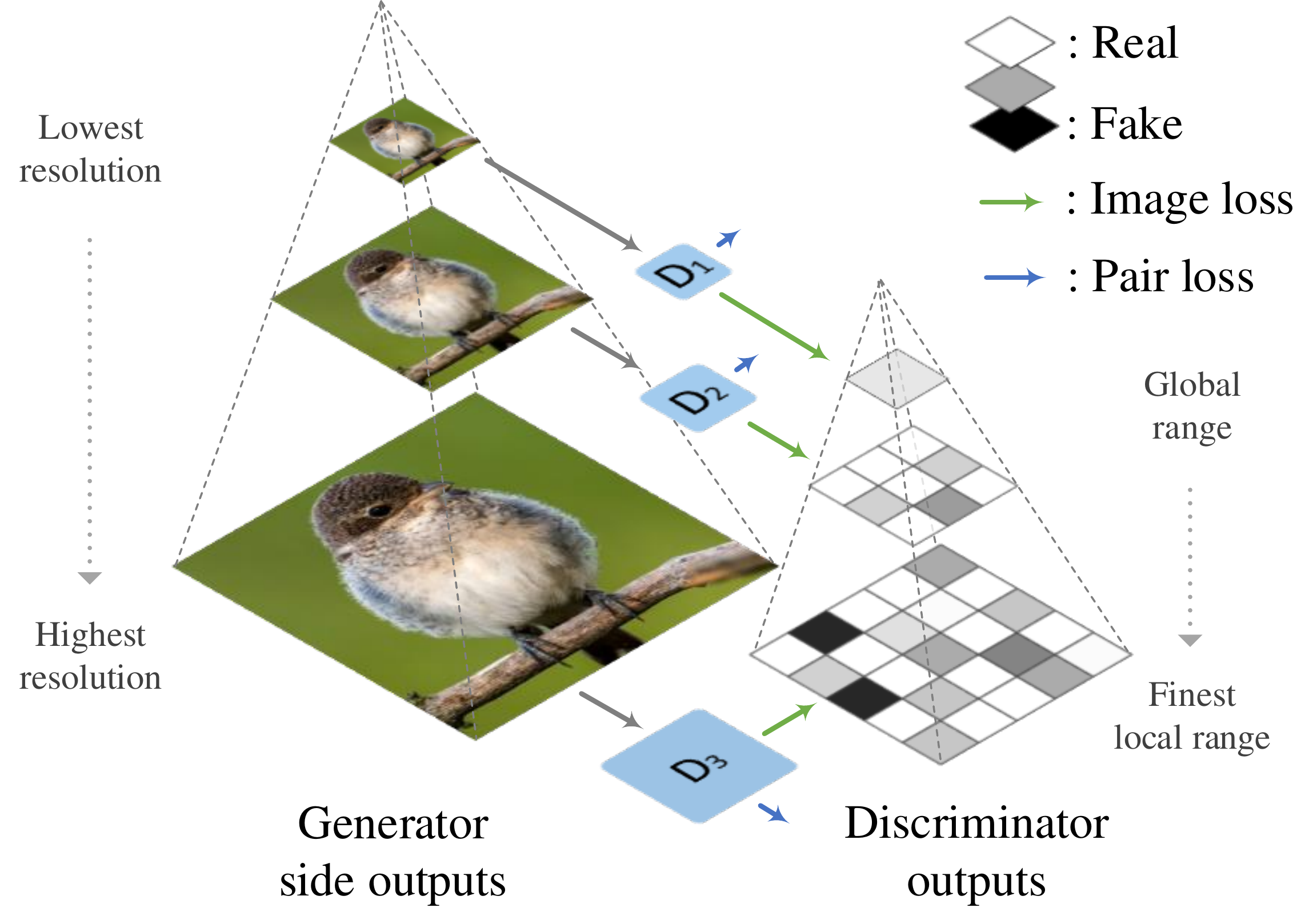}
    \vspace{-.7cm}
    \caption{For each side out image in the pyramid from the generator, the corresponding discriminator $D_i$ computes the matching-aware pair loss and the local image loss (outputting a $R_i{\times}R_i$ probability map to classify real or fake image patches). }  \vspace{-.4cm}
    \label{fig:loss}
\end{figure}

To alleviate these issues and guarantee image fidelity, our solution is to add local adversarial image losses.
We expect the low-resolution discriminators to focus on global structures, while the high-resolution discriminators to focus on local image details.
Each discriminator $D_i$ consists of two branches (see Section 3.4), one computes a single scalar value for the pair loss and another branch computes a $R_i{\times}R_i$ 2D probability map $O_i$ for the local image loss.
For each $D_i$, we control $R_i$ accordingly to tune the receptive field of each element in $O_i$, which 
differentiates whether a corresponding local image patch is real or fake.
The local GAN loss is also used for pixel-to-pixel translation tasks \cite{shrivastava2016learning,zhu2017unpaired,isola2016image}. 
Figure \ref{fig:loss} illustrates how hierarchically-nested discriminators compute the two losses on the generated images in the pyramid. 


\textbf{Full Objective } Overall, our full min-max adversarial objective can be defined as 
\vspace{-.2cm}
\begin{equation}
\begin{split}
\mathcal{V}(& \mathcal{G},\mathcal{D}, \mathcal{Y}, \bm t, \bm z) = \sum_{i=1}^{s} \Big(  L_2\big(D_i({Y}_i)\big) +  L_2\big(D_i({Y}_i, \bm t_{Y})\big) \, + \\ 
& \overline{L_2}\big(D_i({X}_i)\big)  + \overline{L_2}\big(D_i({X}_i, \bm{t}_{X_i})\big) + \overline{L_2}\big(D_i({Y}_i,  \bm{t}_{\overline{Y}})\big) \Big),
\end{split} \vspace{-.6cm}
\end{equation}
where $L_2(x) = \mathbb{E}\big[(x - \mathbb{I})^2\big]$ is the mean-square loss (instead of the conventional cross-entropy loss) and $\overline{L_2}(x) =\mathbb{E}\big[x^2\big]$. 
This objective is minimized by $\mathcal{D}$. While in practice, $\mathcal{G}$ minimizes
$\sum_{i=1}^{s} (L_2(D_i(\mathcal{G}(\bm t, \bm z)_i)) +  L_2\big(D_i(\mathcal{G}(\bm t, \bm z)_i, \bm{t}_{X_i})))$ instead.
For the local image loss, the shape of $x, \mathbb{I} \in \mathbb{R}^{R_i{\times}R_i}$ varies accordingly (see Figure \ref{fig:loss}). $R_i=1$ refers to the (largest local) global range. $D_i(X_i)$ is the image loss branch and $D_i(X_i, \bm{t}_{X_i})$ is the pair loss branch (conditioned on $\bm{t}_{X_i}$).
$\{Y_i, \bm t_{Y}\}$ denotes a matched image-text pair and $\{Y_i, \bm{t}_{\overline{Y}}\}$ denotes a mismatched image-text pair. 

In the spirit of variational auto-encoder \cite{vae} and the practice of StackGAN \cite{han2017stackgan} (termed conditioning augmentation (CA)), instead of directly using the deterministic text embedding, we sample a stochastic vector from a Ganssian distribution $\mathcal{N}(\mu({\bm t}), \Sigma({\bm t}))$, where $\mu$ and $\Sigma$ are parameterized functions of $\bm t$. 
We add the Kullback-Leibler divergence regularization term, $D_{KL}(\mathcal{N}(\mu({\bm t}), \Sigma({\bm t}) )|| \mathcal{N}(0, \bm{I}))$, to the GAN objective to prevent over-fitting and force smooth sampling over the text embedding distribution.

\subsection{Architecture Design}

\textbf{Generator} The generator is simply composed of three kinds of modules, termed $K$-repeat res-blocks, stretching layers, and linear compression layers.
A single res-block in the $K$-repeat res-block is a modified\footnote{We remove ReLU after the skip-addition of each residual block, with an intention to reduce sparse gradients.} residual block \cite{he2016identity}, which contains two convolutional (conv) layers (with batch normalization (BN) \cite{ioffe2015batch} and ReLU). The stretching layer serves to change the feature map size and dimension. It simply contains a scale-$2$ nearest up-sampling layer followed by a conv layer with BN+ReLU. The linear compression layer is a single conv layer followed by a Tanh to directly compress feature maps to the RGB space. We prevent any non-linear function in the compression layer that could impede the gradient signals. 
Starting from a $1024{\times}4{\times}4$ embedding, which is computed by CA and a trained embedding matrix, the generator simply uses $M$ $K$-repeat res-blocks connected by $M{-}1$ in-between stretching layers until the feature maps reach to the targeting resolution. 
For example, for $256{\times}256$ resolution and $K{=}1$, there are $M{=}6$ $1$-repeat res-blocks and $5$ stretching layers. 
At pre-defined side-output positions at scales $\{1,...,s\}$, we apply the compression layer to generate side output images, for the inputs of discriminators. 

\textbf{Discriminator} The discriminator simply contains consecutive stride-2 conv layers with BN+LeakyReLU. There are two branches are added to the upper layer of the discriminator. One branch is a direct fully convolutional layers to produce a $R_i{\times}R_i$ probability map (see Figure \ref{fig:loss}) and classify each location as real or fake. 
Another branch first concatenates a $512{\times}4{\times}4$ feature map and a $128{\times}4{\times}4$ text embedding (replicated by a reduced $128$-d text embedding). Then we use an $1{\times}1$ conv to fuse text and image features and a $4{\times}4$ conv layer to classify an image-text pair to real or fake.

The optimization is similar to the standard alternative training strategy in GANs. Please refer to the supplementary material for more training and network details.

\section{Experiments}
We denote our method as \textbf{HDGAN}, referring to High-Definition results and the idea of Hierarchically-nested Discriminators.

\textbf{Dataset} 
We evaluate our model on three widely used datasets. The CUB dataset \cite{welinder2010caltech} contains 11,788 bird images belonging to 200 categories. 
The Oxford-102 dataset \cite{Nilsback08} contains 8,189 flow images in 102 categories. 
Each image in both datasets is annotated with 10 descriptions provided by \cite{reed2016generative}. We pre-process and split the images of CUB and Oxford-102 following the same pipeline in \cite{reed2016generative,han2017stackgan}. The COCO dataset \cite{lin2014microsoft} contains 82,783 training images and 40,504 validation images. Each image has 5 text annotations. 
We use the pre-trained char-RNN text encoder provided by \cite{reed2016generative} to encode each sentence into a 1024-d text embedding vector.

\textbf{Evaluation metric}
We use three kinds of quantitative metrics to evaluate our method.
1) Inception score \cite{improvedGAN} is a measurement of both objectiveness and diversity of generated images. Evaluating this score needs a pre-trained Inception model~\cite{inception} on ImageNet. For CUB and Oxford-102, we use the fine-tuned Inception models on the training sets of the two datasets, respectively, provided by StackGAN. 
2) Multi-scale structural similarity (MS-SSIM) metric \cite{improvedGAN} is used for further validation. It tests pair-wise similarity of generated images and can identity mode collapses reliably \cite{odena2016conditional}. Lower score indicates higher diversity of generated images (i.e. less model collapses). 

3) \textbf{Visual-semantic similarity} The aforementioned metrics are widely used for evaluating standard GANs. However, they can not measure the alignment between generated images and the conditioned text, i.e.,  semantic consistency. \cite{han2017stackgan} resorts to human evaluation, but this procedure is expensive and difficult to conduct.
To tackle this issue, we introduce a new measurement inspired by \cite{vsemb}, namely visual-semantic similarity (VS similarity). The insight is to train a visual-semantic embedding model and use it to measure the distance between synthesized images and input text. 
Denote $\bm{v}$ as an image feature vector extracted by an Inception model $f_{cnn}$.
We define a scoring function $c(\bm{x}, \bm{y})=\frac{\bm{x}\cdot \bm{y}}{||\bm{x}||_2\cdot ||\bm{y}||_2}$. 
Then, we train two mapping functions $f_v$ and $f_t$, which map both real images and paired text embeddings into a common space in $\mathbb{R}^{512}$, by minimizing the following bi-directional ranking loss:
\begin{equation}\vspace{-.2cm}
\begin{split}
  \sum_{ \bm{v} }\sum_{ \bm{t}_{\overline{v}} }~ &\max( 0, \delta  - c(f_v(\bm{v}), f_t( \bm{t}_{{v}}) )+ c(f_v(\bm{v}), f_t( \bm{t}_{\overline{v}}) )     ) +\\
   \sum_{ \bm{t} }\sum_{ \bm{v}_{\overline{t}} }~ &\max( 0,  \delta  - c(f_t(\bm{t}), f_v( \bm{v}_{{v}} )) +  c(f_t(\bm{t}), f_v( \bm{v}_{\overline{t}} ))  ),
\end{split}
\vspace{-.2cm}
\end{equation}
where $\delta$ is the margin, which is set as 0.2. $\{\bm{v}, \bm{t}\}$ is a ground truth image-text pair, and $\{\bm v, \bm t_{\overline{v}} \}$ and $\{ \bm v_{\overline{t}}, \bm{t} \}$ denote mismatched image-text pairs. In the testing stage, given an text embedding $\bm{t}$, and the generated image $\bm{x}$, the VS score can be calculated as $c(f_{cnn}(\bm{x}), \bm{t})$. Higher score indicates better semantic consistency.

\begin{table}[t] 
    \begin{center}
        \small 
        \begin{tabularx}{.466\textwidth}{c|ccc}

            \specialrule{1.5pt}{0pt}{0pt}  
            \multirow{2}{*}{Method}    & \multicolumn{3}{c}{Dataset}    \\ \cline{2-4}
                                     &     CUB        &    Oxford-102  & COCO             \\ \hline
            GAN-INT-CLS     &    $2.88{\pm}.04$        &     $2.66{\pm}.03$        & $7.88{\pm}.07$     \\
            GAWWN       &        $3.60{\pm}.07$        &     -      &          - \\ 
            StackGAN     &        $3.70{\pm}.04$    &     $3.20{\pm}.01$            &  $8.45{\pm}.03^{\star}$        \\ 
            StackGAN++     &        $3.84{\pm}.06$    &     -            &  -    \\  
            TAC-GAN     &    -        &        $3.45{\pm}.05$        & -    \\    \hline
            HDGAN         &    $\bm{4.15{\pm}.05}$    &    $ \bm{3.45{\pm}.07}$    &  $ \bm{11.86{\pm}.18}$  \\ \hline
        \end{tabularx} 
    \end{center}
    \vspace{-.4cm}
    \begin{tablenotes}
        \small
        \item $^\star$Recently, it updated to ${10.62{\pm}.19}$ in its source code.
    \end{tablenotes} \vspace{-.1cm}
    \caption{The Inception-score comparison on the three datasets. HDGAN outperforms others significantly.} \label{table:score}
\end{table}

\begin{figure*}[t]
    \centering
    \includegraphics[width=0.99\textwidth]{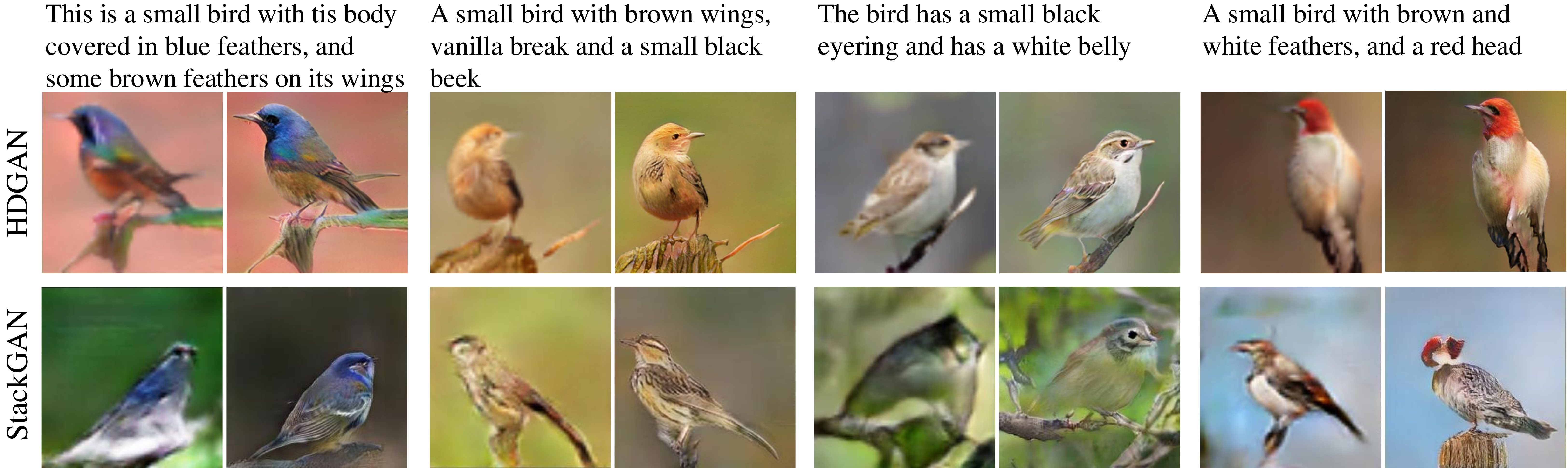}
    \vspace{-.2cm}
    \caption{Generated images on CUB compared with StackGAN. Each sample shows the input text and generated $64^2$ (left) and $256^2$ (right) images. Our results have significantly higher quality and preserve more semantic details, for example, ``the brown and white feathers and read head'' in the last column is much better reflected in our images. Moreover, we observed our birds exhibit nicer poses (e.g. the frontal/back views in the second/forth columns). Zoom-in for better observation.}
    \vspace{-0.1cm}
    \label{fig:vis-cub}
\end{figure*}
\begin{figure*}[t]
    \centering
    \includegraphics[width=0.99\textwidth]{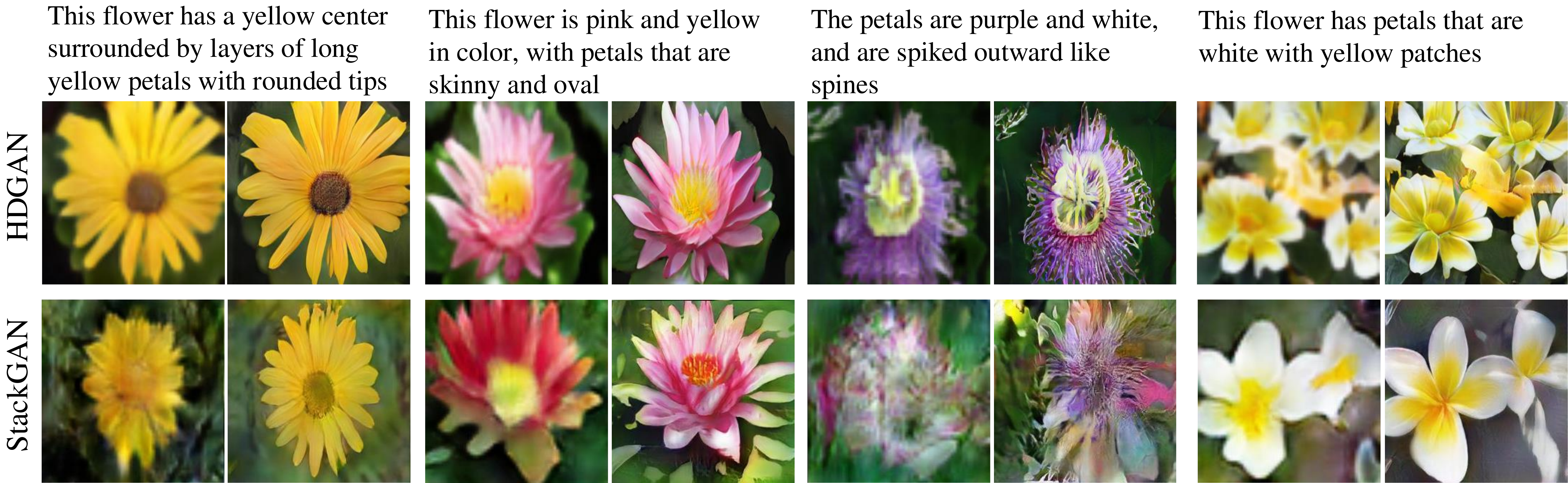}
    \vspace{-.2cm}
    \caption{Generated images on Oxford-102 compared with StackGAN. Our generated images perform more natural satisfiability and higher contrast and can generate complex flower structures (e.g. spiked petals in the third example).} \label{fig:vis-oxford}
    \vspace{-.2cm}
\end{figure*}

\begin{table}[t] 
    \small
    \begin{center}
        \begin{tabularx}{.475\textwidth}{c|ccc}
            \specialrule{1.5pt}{0pt}{0pt}  
            \multirow{2}{*}{Method}    & \multicolumn{3}{c}{Dataset}    \\ \cline{2-4}
            &     CUB        &    Oxford-102  & COCO             \\ \hline
            Ground Truth    &    ${.302{\pm}.151}$    &    $ {.336{\pm}.138}$            & $.426{\pm}.157$  \\ \hline
            StackGAN     &    $.228{\pm}.162$    &     $.278{\pm}.134$            &  $-$        \\ 
            HDGAN         &    $\bm{.246{\pm}.157}$    &    $ \bm{.296{\pm}.131}$ & $\bm{.199{\pm}.183}$  \\ \hline
        \end{tabularx} 
    \end{center}
    \vspace{-.4cm}
    \caption{The VS similarity evaluation on the three datasets. The higher score represents higher semantic consistency between the generated images and conditioned text. The groundtruth score is shown in the first row.} \label{table:vss} \vspace{-.3cm}
\end{table}

\subsection{Comparative Results}
To validate our proposed HDGAN, we compare our results with GAN-INT-CLS \cite{reed2016generative}, GAWWN \cite{reed2016learning}, TAC-GAN \cite{dash2017tac},  Progressive GAN \cite{Karras2017progressive}, StackGAN \cite{han2017stackgan} and also its improved version StackGAN++ \cite{han2017stackganv2}\footnote{StackGAN++ and Prog.GAN are two very recently released preprints we noticed. We acknowledge them as they also target at generating high-resolution images. }. We especially compare with StackGAN in details (results are obtained from its provided models).

Table \ref{table:score} compares the Inception score. We follow the experiment settings of StackGAN to sample ${\sim}30,000$ $256^2$ images for computing the score.
HDGAN achieves significant improvement compared against other methods. For example, it improves StackGAN by $.45$ and StackGAN++ by $.31$ on CUB.
HDGAN achieves competitive results with TAC-GAN on Oxford-102. TAC-GAN uses image labels to increase the discriminability of generators, while we do not use any extra knowledge. 
Figure \ref{fig:vis-cub} and Figure \ref{fig:vis-oxford} compare the qualitative results with StackGAN on CUB and Oxford-102, respectively, by demonstrating more, semantic details, natural color, and complex object structures. 
Moreover, we qualitatively compare the diversity of samples conditioned on the same text (with random input noises) in Figure \ref{fig:multiple-test} left. HDGAN can generate substantially more compelling samples. 

Different from CUB and Oxford-102, COCO is a much more challenging dataset and contains largely diverse natural scenes. 
Our method significantly outperforms StackGAN as well (Table \ref{table:score}). 
Figure \ref{fig:coco} also shows some generated samples in several different scenes. Please refer to the supplementary material for more results.

\begingroup
\setlength{\intextsep}{-4pt}%
\setlength{\columnsep}{0pt}%
\begin{wrapfigure}{r}{0.24\textwidth}
	\centering
	\includegraphics[width=0.245\textwidth]{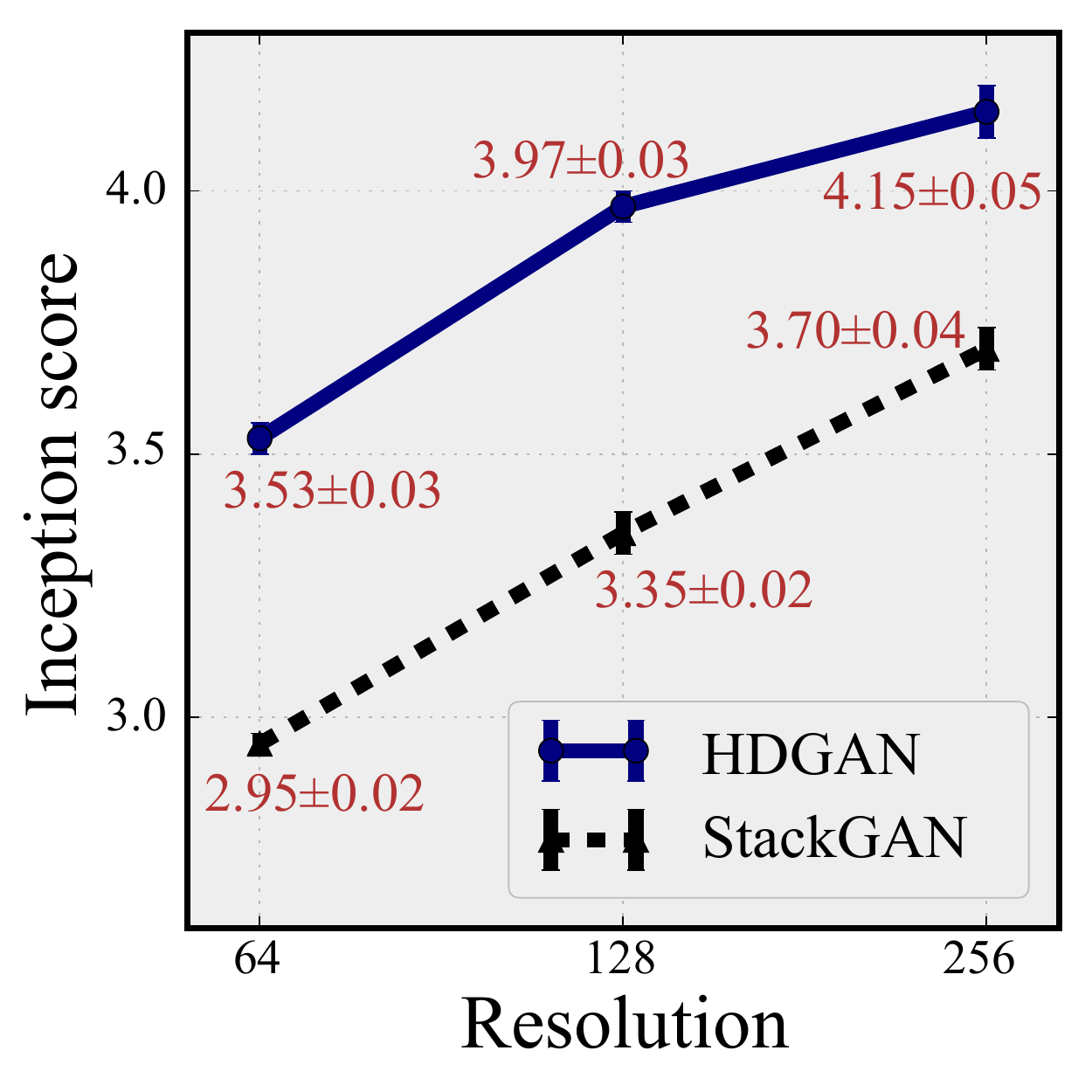}
	\vspace{-12pt}
\end{wrapfigure}
Furthermore, the right figure compares the multi-resolution Inception score on CUB. Our results are from the side outputs of a single model. As can be observed, our $64^2$ results outperform the $128^2$ results of StackGAN and our $128^2$ results also outperform $256^2$ results of StackGAN substantially. It demonstrates that our HDGAN better preserves semantically consistent information in all resolutions (as stated in Section 3.2). Figure \ref{fig:multiple-test} right validates this property qualitatively. On the other hand, we observe that, in StackGAN, the low-resolution images and high-resolution images sometimes are visually inconsistent (see examples in Figure \ref{fig:vis-cub} and \ref{fig:vis-oxford}).

\begin{figure}[t]
	\centering
	\includegraphics[width=0.48\textwidth]{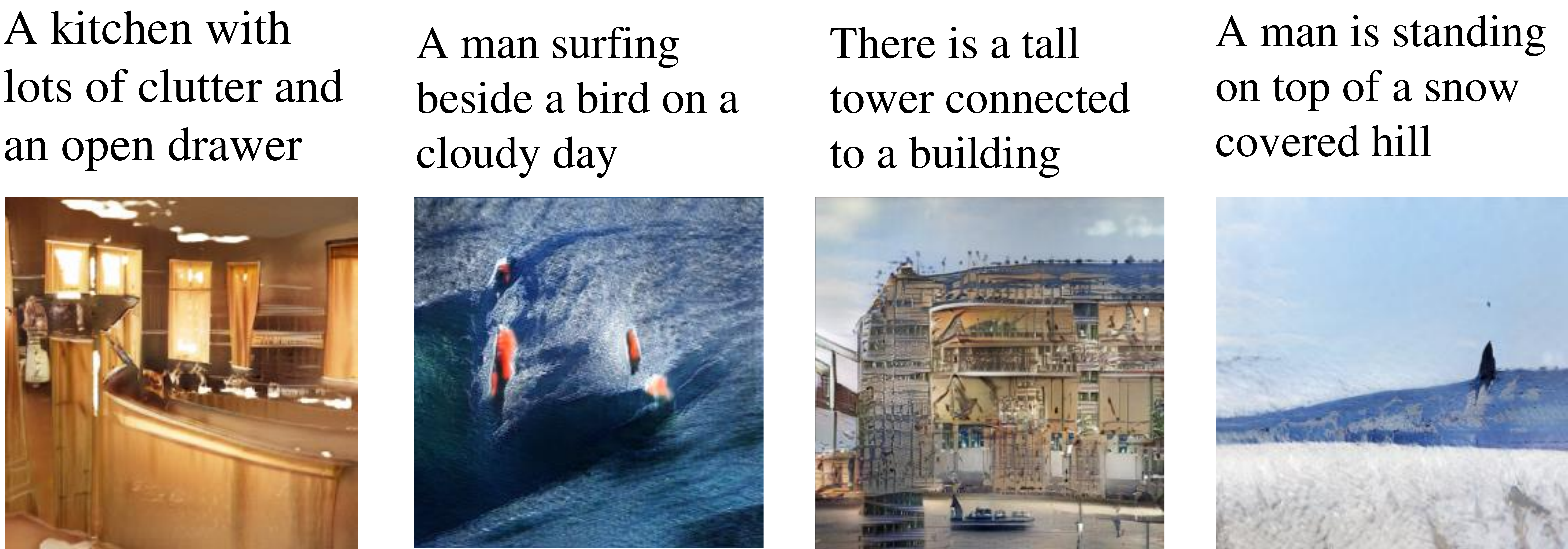}
	\vspace{-.6cm}
	\caption{Samples on the COCO validation set, which contain descriptions across different scenes. } \label{fig:coco}
	\vspace{-.7cm}
\end{figure}

\begin{figure*}[t]
	\centering
	\begin{subfigure}[t]{0.69\textwidth}
		\includegraphics[width=0.99\textwidth]{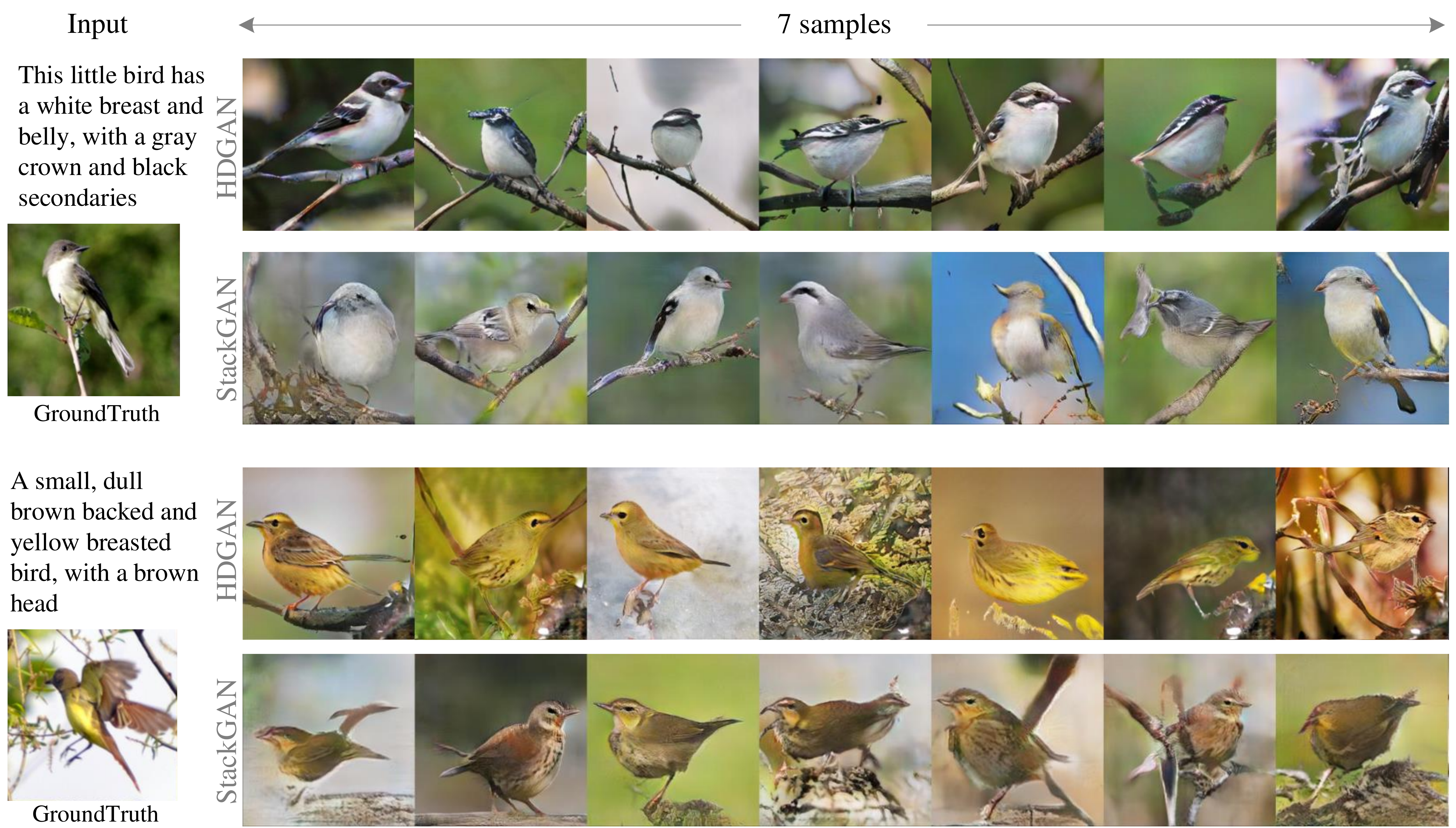}
	\end{subfigure} 
	\begin{subfigure}[t]{0.3\textwidth}
		\includegraphics[width=0.99\textwidth]{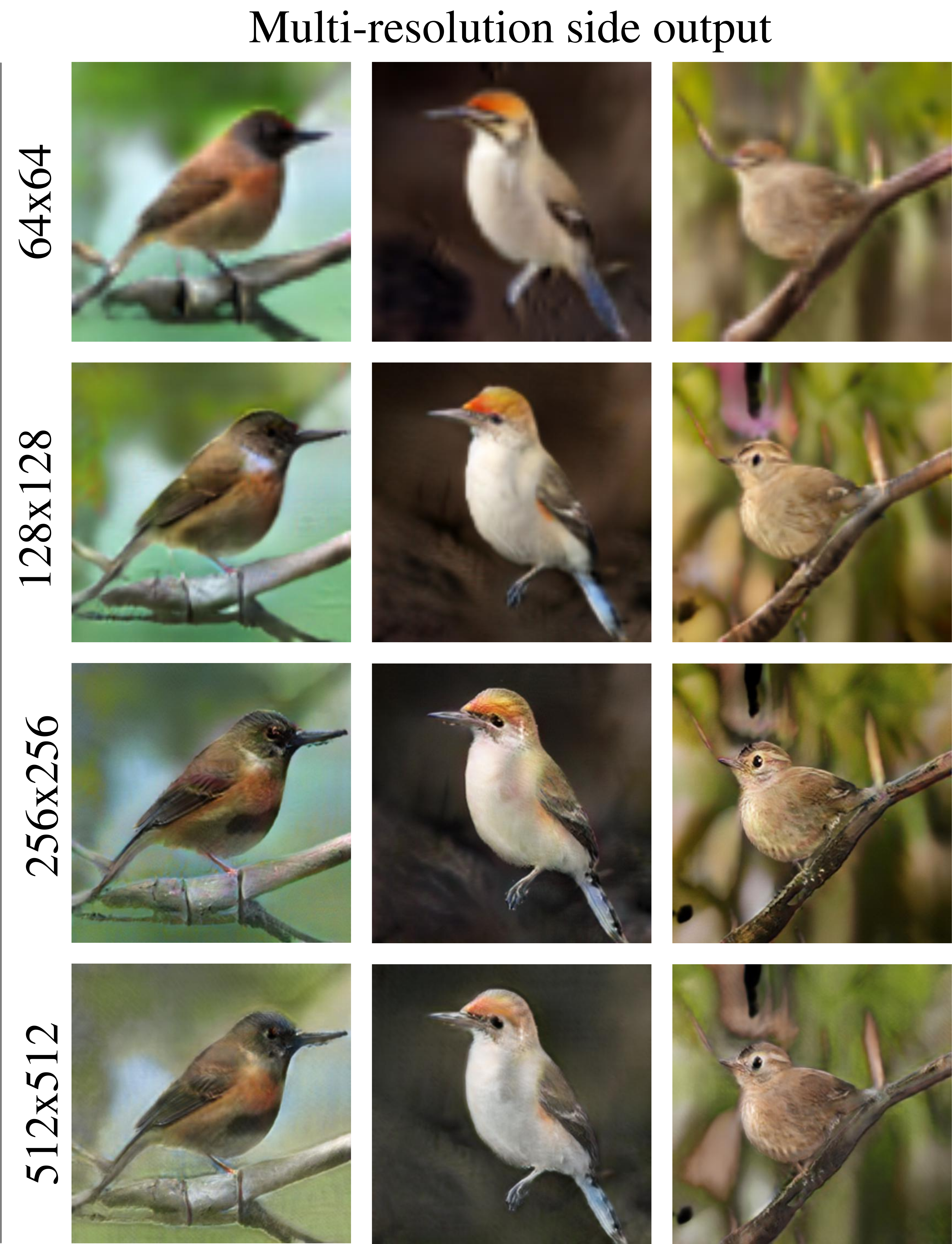}
	\end{subfigure}
	\vspace{-.2cm}
	\caption{Left: Multiple samples are shown given a single input text. The proposed HDGAN (top) show obviously more fine-grained details. Right: Side outputs of HDGAN with increasing resolutions. Different resolutions are semantically consistent and semantic details appear as the resolution increases.  \label{fig:multiple-test}}     \vspace{-.3cm}
\end{figure*}

Table \ref{table:vss} compares the proposed visual-semantic similarity (VS) results on three datasets. The scores of the groundtruth image-text pair are also shown for reference.  HDGAN achieves consistently better performance on both CUB and Oxford-102. These results demonstrate that HDGAN can better capture the visual semantic information in generated images.


Table \ref{fig:msssmi} compares the MS-SSIM score with StackGAN and Prog.GAN for bird image generation. StackGAN and our HDGAN use text as input so the generated images are separable in class. We randomly sample ${\sim}20,000$ image pairs (400 per class) and compare the class-wise score in the left figure. HDGAN outperforms StackGAN in majority of classes and also has a lower standard deviation ($.023$ vs. $.032$).
Note that Prog.GAN uses a noise input rather than text. We can compare with it for a general measure of the image diversity. Following the procedure of Prog.GAN, we randomly sample ${\sim}10,000$ image pairs from all generated samples\footnote{We use $256^2$ bird images provided by Prog.GAN at \url{https://github.com/tkarras/progressive_growing_of_gans}. Note that Prog.GAN is trained on the LSUN \cite{yu2015lsun} bird set, which contains ${\sim}2$ million bird images.} and show the results in Table \ref{fig:msssmi} right. HDGAN outperforms both methods.
 
\begin{table}[t] 
    \begin{minipage}[b]{0.50\linewidth}
        \includegraphics[width=0.99\textwidth,height=0.7\textwidth]{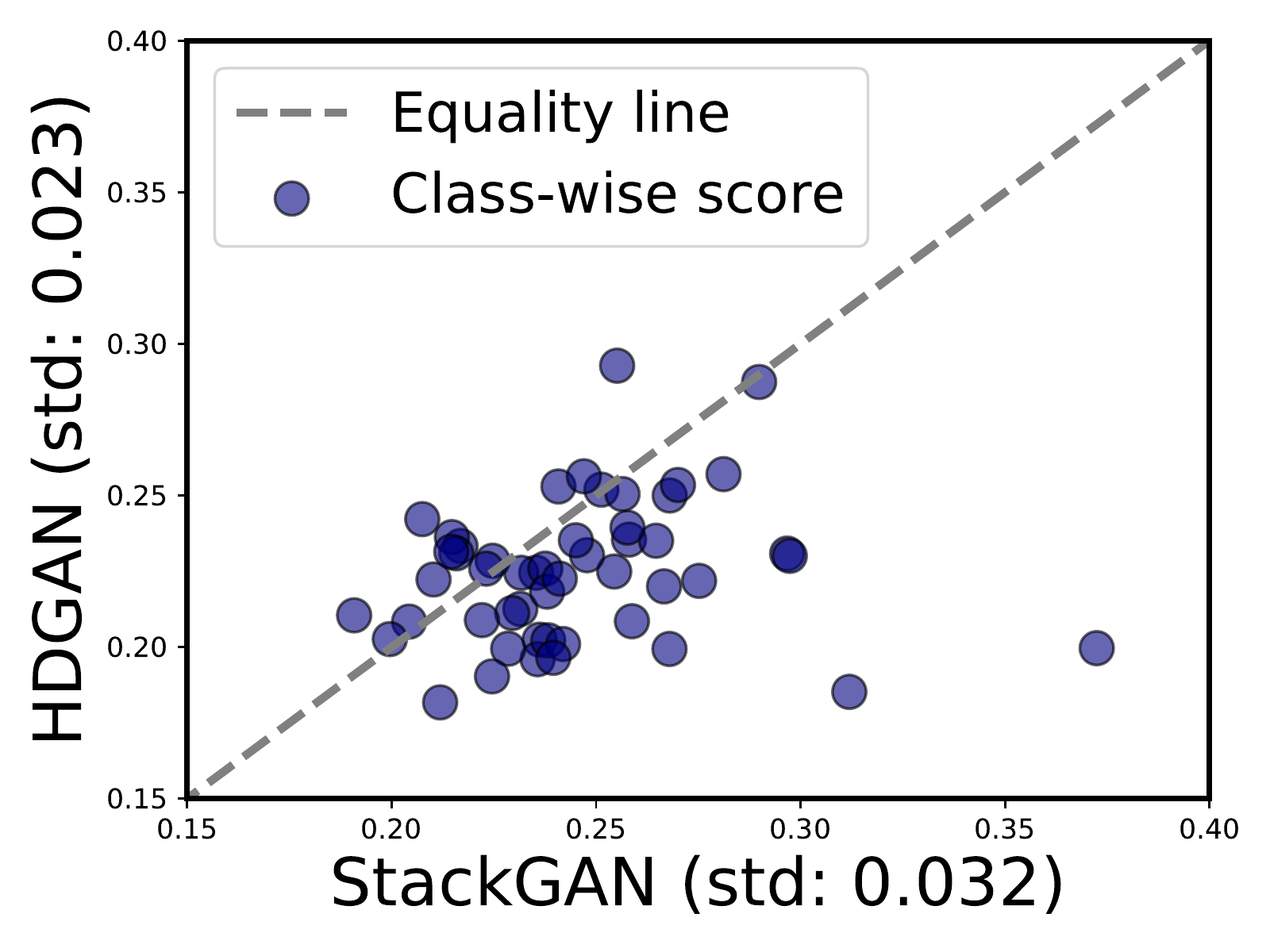}
        \vspace{-1.8cm}
    \end{minipage} 
    \begin{minipage}[b]{0.49\linewidth}
        \begin{tabularx}{.9\textwidth}{c|c}
            \specialrule{1.5pt}{0pt}{0pt}  
            Method   &  MS-SSIM \\ \hline
            StackGAN &   0.234   \\ 
            Prog.GAN &   0.225    \\ \hline
            HDGAN    &   \textbf{0.215}    \\ \hline
        \end{tabularx}
    \end{minipage}
    \vspace{0.2cm}
    \caption{Left: Class-wise MS-SSIM evaluation. Lower score indicates higher intraclass dissimilarity. The points below the equality line represent classes our HDGAN wins. The inter-class std is shown in axis text. Right: Overall (not class-wised)  MS-SSIM evaluation.} \label{fig:msssmi}
    \vspace{-0.4cm}
\end{table}

\subsection{Style Transfer Using Sentence Interpolation}
Ideally, a well-trained model can generalize to a smooth linear latent data manifold. To demonstrate this capability, we generate images using the linearly interpolated embeddings between two source sentences. 
As shown in Figure~\ref{fig:interp}, our generated images show smooth style transformation and well reflect the semantic details in sentences. 
For example, in the second row, complicated sentences with detailed appearance descriptions (e.g. pointy peaks and black wings) are used, our model could still successfully capture these subtle features and tune the bird's appearance smoothly. 



\subsection{Ablation Study and Discussion}
\textbf{Hierarchically-nested adversarial training} Our hierarchically-nested discriminators play a role of regularizing the layer representations (at scale $\{64, 128, 256\}$). 
In Table \ref{table:multiscales}, we demonstrate their effectiveness and show the performance by removing parts of discriminators on both CUB and COCO datasets. 
As can be seen, increasing the usage of discriminators at different scales have positive effects. And using discriminators at $64^2$ is critical (by comparing the 64-256 and 128-256 cases). For now, it is uncertain if adding more discriminators and even on lower resolutions would be helpful. Further validation will be conducted.
StackGAN emphasizes the importance of using text embeddings not only at the input but also with intermediate features of the generator, by showing a large drop from $3.7$ to $3.45$ without doing so. While our method only uses text embeddings at the input. Our results strongly demonstrate the effectiveness of our hierarchically-nested adversarial training to maintain such semantic information and a high Inception score. 

\begin{figure}[t]
    \centering
    \includegraphics[width=0.48\textwidth]{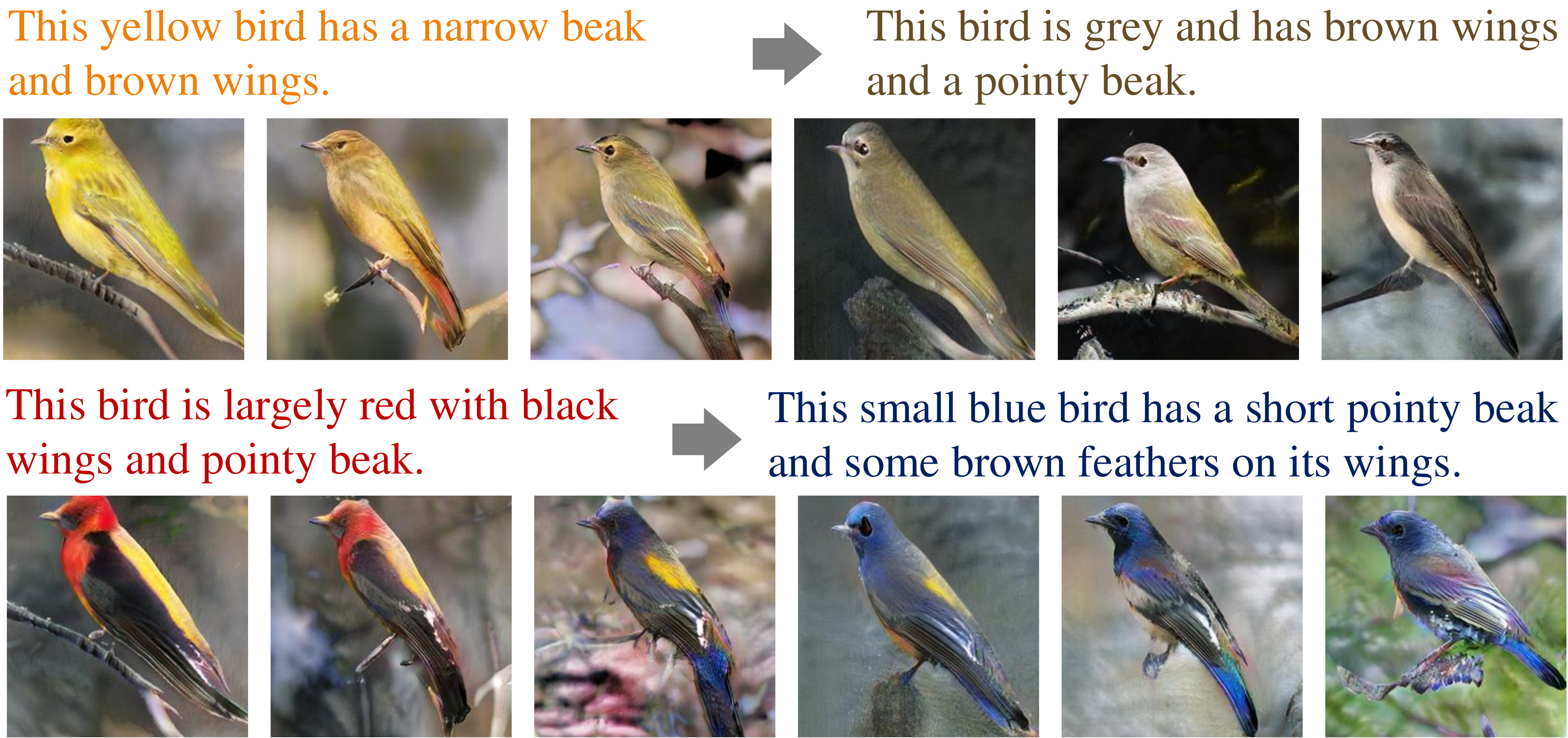}
    \vspace{-.6cm}
    \caption{Text embedding interpolation from the source to target sentence results in smooth image style changes to match the targeting sentence. } \label{fig:interp}
    \vspace{-.3cm}
\end{figure}

%
\begin{table}[t] 
    \begin{center}
        \begin{tabularx}{.36\textwidth}{ccc|cc}
            \specialrule{1.5pt}{0pt}{0pt}  
            \multicolumn{3}{c|}{Discriminators}    &  \multicolumn{2}{c}{Inception score}    \\ \hline
            64    & 128    & 256             &     CUB        &     COCO \\ \hline
            &          &    \checkmark    &    ${3.52{\pm}.04}$& -    \\ 
            &      \checkmark    &    \checkmark    &    ${3.99{\pm}.04}$&-    \\
            \checkmark    &              &    \checkmark    &  ${4.14{\pm}.03}$    & ${11.29{\pm}.18}$    \\
            \checkmark    &  \checkmark        &    \checkmark    &    ${4.15{\pm}.05}$& ${11.86{\pm}.18}$ \\ \hline
            
        \end{tabularx}
    \end{center} \vspace{-.5cm}
    \caption{Ablation study of hierarchically-nested adversarial discriminators on CUB and COO. $\checkmark $ indicates whether a discriminator at a certain scale is used. See text for detailed explanations.} \label{table:multiscales} \vspace{-.3cm}
\end{table}

%



\textbf{The local image loss} We analyze the effectiveness of the proposed local adversarial image loss. 
Table \ref{table:multiscales} compares the case without using it (denoted as `w/o local image loss').
The local image loss helps improve the visual-semantic matching evidenced by a higher VS score. We hypothesize that it is because adding the separate local image loss can offer the pair loss more ``focus'' on learning the semantic consistency. 
Furthermore, the local image loss helps generate more vivid image details. As demonstrated in Figure \ref{fig:vis-imgloss}, although both models can successfully capture the semantic details in the text, the `w/ local' model generates complex object structures described in the conditioned text more precisely.

\textbf{Design principles} StackGAN claims the failure of directly training a vanilla $256^2$ GAN to generate meaningful images. 
We test this extreme case using our method by removing all nested discriminators without the last one. Our method still generates fairly meaningful results (the first row of Table \ref{table:multiscales}), which demonstrate the effectiveness of our proposed framework (see Section 3.4).

Initially, we tried to share the top layers of the hierarchical-nested discriminators of HDGAN inspired by \cite{liu2017unsupervised}. The intuition is that all discriminators have a common goal to differentiate real and fake despite difficult scales, and such sharing would reduce their inter-variances. 
However, we did not observe benefits from this mechanism and our independent discriminators can be trained fairly stably. 

HDGAN has a very succinct framework, compared most existing methods, as they \cite{xu2017attngan,char2017perceptual} adds extra supervision on output images to `inject' semantic information, which is shown helpful for improving the inception score. However, it is not clear that whether these strategies can substantially improve the visual quality, which is worth further study.

%

\section{Conclusion}
In this paper, we present a novel and effective method to tackle the problem of generating images conditioned on text descriptions. We explore a new dimension of playing adversarial games along the depth of the generator using the hierarchical-nested adversarial objectives. A multi-purpose adversarial loss is adopted to help render fine-grained image details.
We also introduce a new evaluation metric to evaluate the semantic consistency between generated images and conditioned text.
Extensive experiment results demonstrate that our method, namely HDGAN, can generate high-resolution photographic images and performs significantly better than existing state of the arts on three public datasets.
\begin{table}[t] 
    \small
    \centering
    \begin{tabularx}{0.37\textwidth}{c|c|c}
        \specialrule{1.5pt}{0pt}{0pt}  
        &   Inc. score      & VS       \\  \hline
        w/o  local image loss      &   $3.12{\pm}.02$      & $.263{\pm}.130$           \\  \hline
        w/  local  image loss      &   $3.45{\pm}.07$      &   $.296{\pm}.130$            \\ \hline
    \end{tabularx}
    \vspace{-0.2cm}
    \caption{Ablation study of the local image loss on Oxford-102. See text for detailed explanations.} \label{tab:ablation-loss} \vspace{-0.3cm}
\end{table}
\begin{figure}[t]
    \centering
    \includegraphics[width=0.491\textwidth,height=0.29\textwidth]{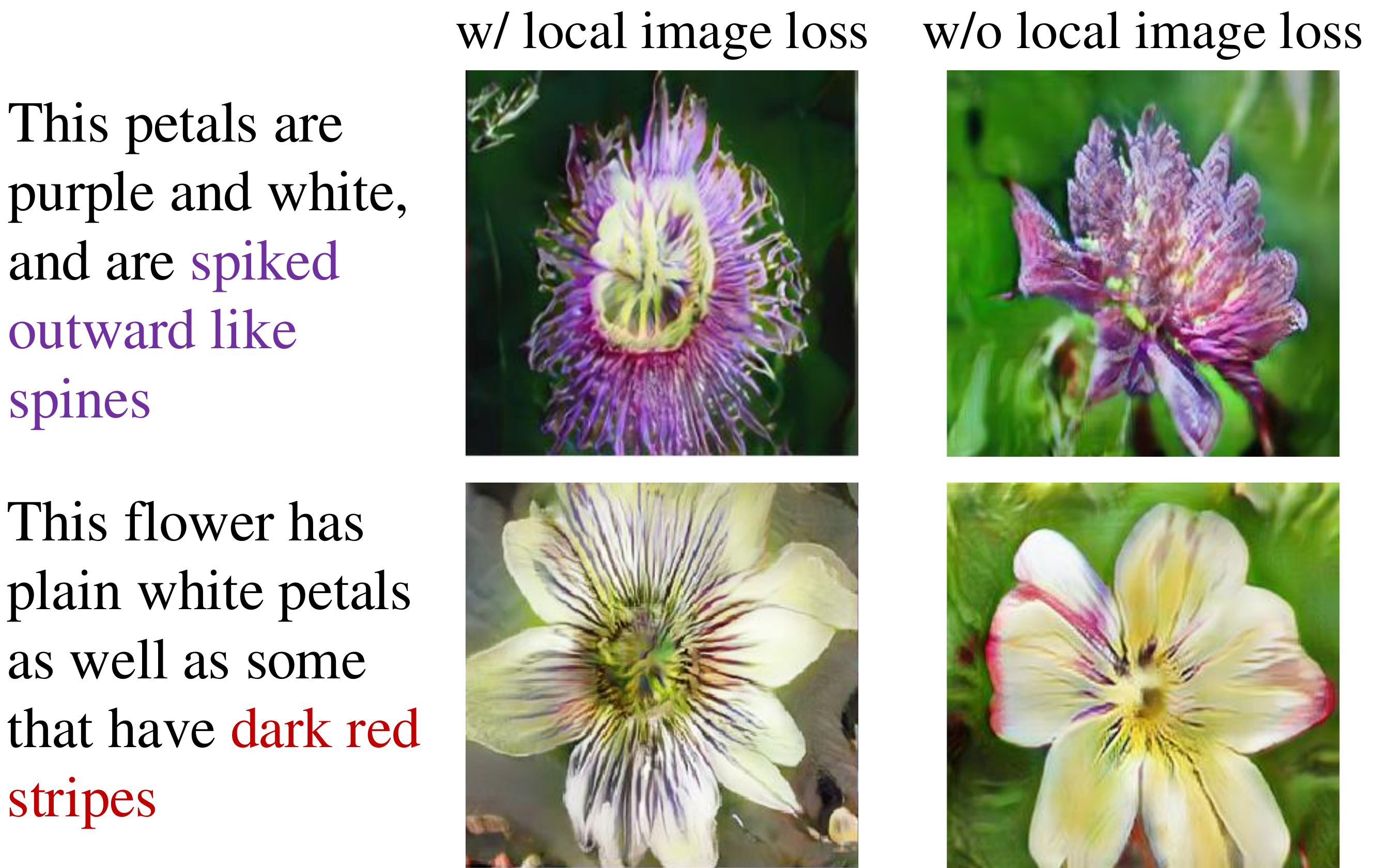}
    \vspace{-.6cm}
    \caption{Qualitative evaluation of the local image loss. The two images w/ the local image loss more precisely exhibit the complex flower petal structures described in the (colored) text.} \label{fig:vis-imgloss}
    \vspace{-.2cm}
\end{figure}

{\small
\bibliographystyle{ieee}
\bibliography{reference_zizhao,egbib}

\begin{thebibliography}{10}\itemsep=-1pt

\bibitem{ba2016layer}
J.~L. Ba, J.~R. Kiros, and G.~E. Hinton.
\newblock Layer normalization.
\newblock {\em arXiv preprint arXiv:1607.06450}, 2016.

\bibitem{berthelot2017began}
D.~Berthelot, T.~Schumm, and L.~Metz.
\newblock Began: Boundary equilibrium generative adversarial networks.
\newblock {\em arXiv preprint arXiv:1703.10717}, 2017.

\bibitem{char2017perceptual}
M.~Cha, Y.~Gwon, and H.~T. Kung.
\newblock Adversarial nets with perceptual losses for text-to-image synthesis.
\newblock {\em arXiv preprint arXiv:1708.09321}, 2017.

\bibitem{chen2017photographic}
Q.~Chen and V.~Koltun.
\newblock Photographic image synthesis with cascaded refinement networks.
\newblock {\em ICCV}, 2017.

\bibitem{costa2017towards}
P.~Costa, A.~Galdran, M.~I. Meyer, M.~D. Abr{\`a}moff, M.~Niemeijer, A.~M.
  Mendon{\c{c}}a, and A.~Campilho.
\newblock Towards adversarial retinal image synthesis.
\newblock {\em arXiv preprint arXiv:1701.08974}, 2017.

\bibitem{dash2017tac}
A.~Dash, J.~C.~B. Gamboa, S.~Ahmed, M.~Z. Afzal, and M.~Liwicki.
\newblock Tac-gan-text conditioned auxiliary classifier generative adversarial
  network.
\newblock {\em arXiv preprint arXiv:1703.06412}, 2017.

\bibitem{denton2015deep}
E.~L. Denton, S.~Chintala, R.~Fergus, et~al.
\newblock Deep generative image models using a￼ laplacian pyramid of
  adversarial networks.
\newblock In {\em NIPS}, pages 1486--1494, 2015.

\bibitem{dong2017semantic}
H.~Dong, S.~Yu, C.~Wu, and Y.~Guo.
\newblock Semantic image synthesis via adversarial learning.
\newblock {\em ICCV}, 2017.

\bibitem{dosovitskiy2016generating}
A.~Dosovitskiy and T.~Brox.
\newblock Generating images with perceptual similarity metrics based on deep
  networks.
\newblock In {\em NIPS}, 2016.

\bibitem{durugkar2016generative}
I.~Durugkar, I.~Gemp, and S.~Mahadevan.
\newblock Generative multi-adversarial networks.
\newblock {\em arXiv preprint arXiv:1611.01673}, 2016.

\bibitem{goodfellow2014generative}
I.~Goodfellow, J.~Pouget-Abadie, M.~Mirza, B.~Xu, D.~Warde-Farley, S.~Ozair,
  A.~Courville, and Y.~Bengio.
\newblock Generative adversarial nets.
\newblock In {\em NIPS}, 2014.

\bibitem{he2016identity}
K.~He, X.~Zhang, S.~Ren, and J.~Sun.
\newblock Identity mappings in deep residual networks.
\newblock In {\em ECCV}, 2016.

\bibitem{huang2016stacked}
X.~Huang, Y.~Li, O.~Poursaeed, J.~Hopcroft, and S.~Belongie.
\newblock Stacked generative adversarial networks.
\newblock {\em CVPR}, 2017.

\bibitem{ioffe2015batch}
S.~Ioffe and C.~Szegedy.
\newblock Batch normalization: Accelerating deep network training by reducing
  internal covariate shift.
\newblock In {\em ICML}, 2015.

\bibitem{isola2016image}
P.~Isola, J.-Y. Zhu, T.~Zhou, and A.~A. Efros.
\newblock Image-to-image translation with conditional adversarial networks.
\newblock {\em CVPR}, 2017.

\bibitem{johnson2016perceptual}
J.~Johnson, A.~Alahi, and L.~Fei-Fei.
\newblock Perceptual losses for real-time style transfer and super-resolution.
\newblock In {\em ECCV}, 2016.

\bibitem{Karras2017progressive}
T.~Karras, T.~Aila, S.~Laine, and J.~Lehtinen.
\newblock Progressive growing of gans for improved quality, stability, and
  variation.
\newblock In {\em arXiv preprint arXiv:1710.10196}, 2016.

\bibitem{kingma2014adam}
D.~Kingma and J.~Ba.
\newblock Adam: A method for stochastic optimization.
\newblock {\em arXiv preprint arXiv:1412.6980}, 2014.

\bibitem{kingma2013auto}
D.~P. Kingma and M.~Welling.
\newblock Auto-encoding variational bayes.
\newblock {\em arXiv preprint arXiv:1312.6114}, 2013.

\bibitem{vae}
D.~P. Kingma and M.~Welling.
\newblock Auto-encoding variational bayes.
\newblock {\em ICLR}, 2013.

\bibitem{vsemb}
R.~Kiros, R.~Salakhutdinov, and R.~S. Zemel.
\newblock Unifying visual-semantic embeddings with multimodal neural language
  models.
\newblock {\em arXiv preprint arXiv:1411.2539}, 2014.

\bibitem{ledig2016photo}
C.~Ledig, L.~Theis, F.~Husz{\'a}r, J.~Caballero, A.~Cunningham, A.~Acosta,
  A.~Aitken, A.~Tejani, J.~Totz, Z.~Wang, et~al.
\newblock Photo-realistic single image super-resolution using a generative
  adversarial network.
\newblock {\em CVPR}, 2017.

\bibitem{lee2015deeply}
C.-Y. Lee, S.~Xie, P.~Gallagher, Z.~Zhang, and Z.~Tu.
\newblock Deeply-supervised nets.
\newblock In {\em AIS}, 2015.

\bibitem{lin2014microsoft}
T.-Y. Lin, M.~Maire, S.~Belongie, J.~Hays, P.~Perona, D.~Ramanan,
  P.~Doll{\'a}r, and C.~L. Zitnick.
\newblock Microsoft coco: Common objects in context.
\newblock In {\em ECCV}, 2014.

\bibitem{liu2017unsupervised}
M.-Y. Liu, T.~Breuel, and J.~Kautz.
\newblock Unsupervised image-to-image translation networks.
\newblock {\em arXiv preprint arXiv:1703.00848}, 2017.

\bibitem{long2015fully}
J.~Long, E.~Shelhamer, and T.~Darrell.
\newblock Fully convolutional networks for semantic segmentation.
\newblock In {\em CVPR}, 2015.

\bibitem{lsgan}
X.~Mao, Q.~Li, H.~Xie, R.~Y. Lau, Z.~Wang, and S.~P. Smolley.
\newblock Least squares generative adversarial networks.
\newblock {\em arXiv preprint ArXiv:1611.04076}, 2016.

\bibitem{metz2016unrolled}
L.~Metz, B.~Poole, D.~Pfau, and J.~Sohl-Dickstein.
\newblock Unrolled generative adversarial networks.
\newblock {\em arXiv preprint arXiv:1611.02163}, 2016.

\bibitem{tu_etal_nips17_d2gan}
T.~D. Nguyen, T.~Le, H.~Vu, and D.~Phung.
\newblock Dual discriminator generative adversarial nets.
\newblock In {\em NIPS}, 2017.

\bibitem{Nilsback08}
M.-E. Nilsback and A.~Zisserman.
\newblock Automated flower classification over a large number of classes.
\newblock In {\em ICCVGIP}, 2008.

\bibitem{odena2016conditional}
A.~Odena, C.~Olah, and J.~Shlens.
\newblock Conditional image synthesis with auxiliary classifier gans.
\newblock {\em arXiv preprint arXiv:1610.09585}, 2016.

\bibitem{oord2016pixel}
A.~v.~d. Oord, N.~Kalchbrenner, and K.~Kavukcuoglu.
\newblock Pixel recurrent neural networks.
\newblock {\em ICML}, 2016.

\bibitem{radford2015unsupervised}
A.~Radford, L.~Metz, and S.~Chintala.
\newblock Unsupervised representation learning with deep convolutional
  generative adversarial networks.
\newblock {\em arXiv preprint arXiv:1511.06434}, 2015.

\bibitem{reed2016generative}
S.~Reed, Z.~Akata, X.~Yan, L.~Logeswaran, B.~Schiele, and H.~Lee.
\newblock Generative adversarial text to image synthesis.
\newblock {\em ICML}, 2016.

\bibitem{reed2016learning}
S.~E. Reed, Z.~Akata, S.~Mohan, S.~Tenka, B.~Schiele, and H.~Lee.
\newblock Learning what and where to draw.
\newblock In {\em NIPS}, 2016.

\bibitem{salimans2016improved}
T.~Salimans, I.~Goodfellow, W.~Zaremba, V.~Cheung, A.~Radford, and X.~Chen.
\newblock Improved techniques for training gans.
\newblock In {\em NIPS}, 2016.

\bibitem{improvedGAN}
T.~Salimans, I.~Goodfellow, W.~Zaremba, V.~Cheung, A.~Radford, X.~Chen, and
  X.~Chen.
\newblock Improved techniques for training gans.
\newblock In {\em NIPS}. 2016.

\bibitem{shrivastava2016learning}
A.~Shrivastava, T.~Pfister, O.~Tuzel, J.~Susskind, W.~Wang, and R.~Webb.
\newblock Learning from simulated and unsupervised images through adversarial
  training.
\newblock {\em CVPR}, 2017.

\bibitem{inception}
C.~Szegedy, W.~Liu, Y.~Jia, P.~Sermanet, S.~Reed, D.~Anguelov, D.~Erhan,
  V.~Vanhoucke, and A.~Rabinovich.
\newblock Going deeper with convolutions.
\newblock In {\em CVPR}, June 2015.

\bibitem{ulyanov2016instance}
D.~Ulyanov, A.~Vedaldi, and V.~Lempitsky.
\newblock Instance normalization: The missing ingredient for fast stylization.
\newblock {\em arXiv preprint arXiv:1607.08022}, 2016.

\bibitem{welinder2010caltech}
P.~Welinder, S.~Branson, T.~Mita, C.~Wah, F.~Schroff, S.~Belongie, and
  P.~Perona.
\newblock Caltech-ucsd birds 200.
\newblock 2010.

\bibitem{xie2015holistically}
S.~Xie and Z.~Tu.
\newblock Holistically-nested edge detection.
\newblock In {\em ICCV}, 2015.

\bibitem{xu2017attngan}
T.~Xu, P.~Zhang, Q.~Huang, H.~Zhang, Z.~Gan, X.~Huang, and X.~He.
\newblock Attngan: Fine-grained text to image generation with attentional
  generative adversarial networks.
\newblock {\em arXiv preprint arXiv:1711.10485}, 2017.

\bibitem{yu2015lsun}
F.~Yu, A.~Seff, Y.~Zhang, S.~Song, T.~Funkhouser, and J.~Xiao.
\newblock Lsun: Construction of a large-scale image dataset using deep learning
  with humans in the loop.
\newblock {\em arXiv preprint arXiv:1506.03365}, 2015.

\bibitem{zhang2017multistyle}
H.~Zhang and K.~Dana.
\newblock Multi-style generative network for real-time transfer.
\newblock {\em arXiv preprint arXiv:1703.06953}, 2017.

\bibitem{zhang2017image}
H.~Zhang, V.~Sindagi, and V.~M. Patel.
\newblock Image de-raining using a conditional generative adversarial network.
\newblock {\em arXiv preprint arXiv:1701.05957}, 2017.

\bibitem{han2017stackgan}
H.~Zhang, T.~Xu, H.~Li, S.~Zhang, X.~Wang, X.~Huang, and D.~Metaxas.
\newblock Stackgan: Text to photo-realistic image synthesis with stacked
  generative adversarial networks.
\newblock In {\em ICCV}, 2017.

\bibitem{han2017stackganv2}
H.~Zhang, T.~Xu, H.~Li, S.~Zhang, X.~Wang, X.~Huang, and D.~Metaxas.
\newblock Stackgan++: Text to photo-realistic image synthesis with stacked
  generative adversarial networks.
\newblock In {\em arXiv preprint arXiv:1710.10916}, 2017.

\bibitem{Zhang_2017_CVPR}
Z.~Zhang, Y.~Xie, F.~Xing, M.~Mcgough, and L.~Yang.
\newblock Mdnet: A semantically and visually interpretable medical image
  diagnosis network.
\newblock In {\em CVPR}, 2017.

\bibitem{zhang2018cardiac}
Z.~Zhang, L.~Yang, and Y.~Zheng.
\newblock Translating and segmenting multimodal medical volumes with cycle- and
  shape-consistency generative adversarial network.
\newblock In {\em CVPR}, 2018.

\bibitem{zhao2017pyramid}
H.~Zhao, J.~Shi, X.~Qi, X.~Wang, and J.~Jia.
\newblock Pyramid scene parsing network.
\newblock In {\em CVPR}, 2017.

\bibitem{zhu2017unpaired}
J.-Y. Zhu, T.~Park, P.~Isola, and A.~A. Efros.
\newblock Unpaired image-to-image translation using cycle-consistent
  adversarial networks.
\newblock {\em ICCV}, 2017.

\end{thebibliography}
}

\vspace{2.cm}

\section{Supplementary Material}
\subsection{Training and Architecture Details}
The training procedure is similar to the one used in standard GANs, which alternatively updates the generator and discriminators until converge.

The Adam optimizer \cite{kingma2014adam} is used.  The initial learning rate is set as 0.0002 and decreased by half for every 100 epochs (50 for COCO). The model is trained for 500 epochs in total (200 epochs for COCO).
We configure the side outputs at 4 different scales where the feature map resolution is equal to $64^2,128^2,256^2$, and $512^2$, respectively.
For the local image loss of these 4 side outputs, we set $R_1=1, R_2=1, R_3=5, \text{and } R_4=5$. For example, $R_1$ refers to $64^2$. These numbers are not fine-tuned but are set empirically. We believe there exists better configurations to be explored.

All intermediate conv layers, except from the specified ones in Section 3.4, use $3{\times}3$ kernels (with reflection padding).
Some other normalization also layers are experimented (i.e. instance normalization \cite{ulyanov2016instance} and layer normalization \cite{ba2016layer}) since they are used by recent advances \cite{zhu2017unpaired,chen2017photographic}. But the results are not satisfactory. 

With respect to the generator, we use 1-repeat residual blocks for the generator till the $256^2$ resolution. The input of the generator is a $1024{\times}4{\times}4$ tensor.  As the feature map resolution increases by 2, the number of feature maps is halved at $8, 32, 128, 256$ sizes. 
To generate $512^2$ images, we pre-train the generator to $256^2$ due to the limitation of the GPU memory. We use a $3$-repeat res-block followed by a stretching layer to upscale the feature map size to $32{\times}512{\times}512$. and a linear compression layer to generate images. 
Since the $256^2$ image already captures the overall semantics and details, to boost the training and encourage the $512^2$ maintain this information, we use a l1 reconstruction loss to `self-regularize' the generator. 

\subsection{More Qualitative Results and Analysis}
In this section, we demonstrate more sample results for the three datasets.

Figure \ref{fig:bird2} compares our results with StackGAN. For each input, 6 images are randomly sampled. Furthermore, we visualize zoomed-in samples compared with StackGAN in Figure \ref{fig:bird3}. Our results demonstrate obviously better quality, less artifacts, and less sharp pixel transitions.

Figure \ref{fig:bird} shows the results on the CUB bird dataset. All the outputs of a model with different resolutions are also shown. As can be observed in this two figures, our method can generate fairly vivid images with different poses, shape, background, etc. Moreover, the images with different resolutions, which are side outputs of a single model, have very consistent information. More and more image details can be observed as the resolution increases.
Figure \ref{fig:flower} shows the results on the Oxford-102 flower dataset. Very detailed petals can be generated with photographic colors and saturability.

Figure \ref{fig:coco} shows some sampled results on the COCO dataset. COCO is much more challenging than the other two datasets since it contains natural images from a wide variety of scenes containing hundreds of different objects. 
As can be observed in the shown samples, our method can still generate semantically consistent images. 

However, it is worth to notice that, although our method significantly improves existing methods \cite{han2017stackgan,reed2016generative} on COCO, we realize that generating fine-grained details of complex natural scenes with various objects is still challenging. Based on this study, we expect to further address this problem as the future study. 

\textbf{Failure cases}: Although we observed that the majority of test data can result in successful outputs (at least one sample of a single input text), there are still observable failure cases. The major problems include obvious artifacts, minor semantic inconsistency (compared with groundtruth), loss of object basic shapes.
Figure \ref{fig:fail} shows these mentioned failure cases. To compare with StackGAN category by category, please refer to Table 3 left (in the main paper).
\newpage

\begin{figure*}[th!]
	\centering
	\includegraphics[width=0.83\textwidth]{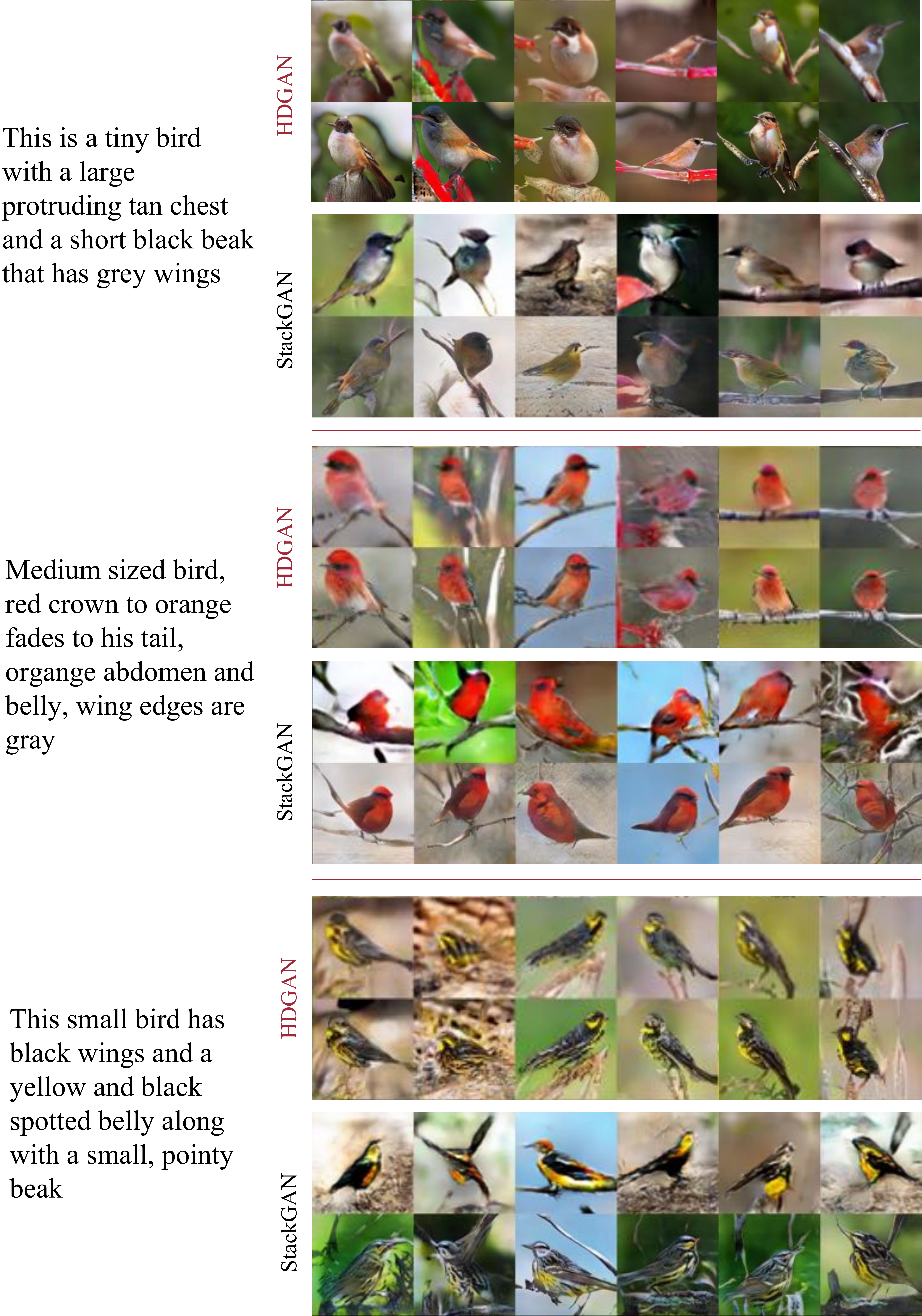}
	
	\caption{Sample results on CUB compared with StackGAN. For each input, 6 samples are shown with resolutions of $64^2$ and $256^2$, respectively. As can be obviously seen, our HDGAN can generate very consistent content in images of different resolutions. Moreover, our generated images show more photographic color and contrast.  }  
	\label{fig:bird2}
\end{figure*}
\newpage
\begin{figure*}[th!]
	\centering
	\includegraphics[width=0.99\textwidth]{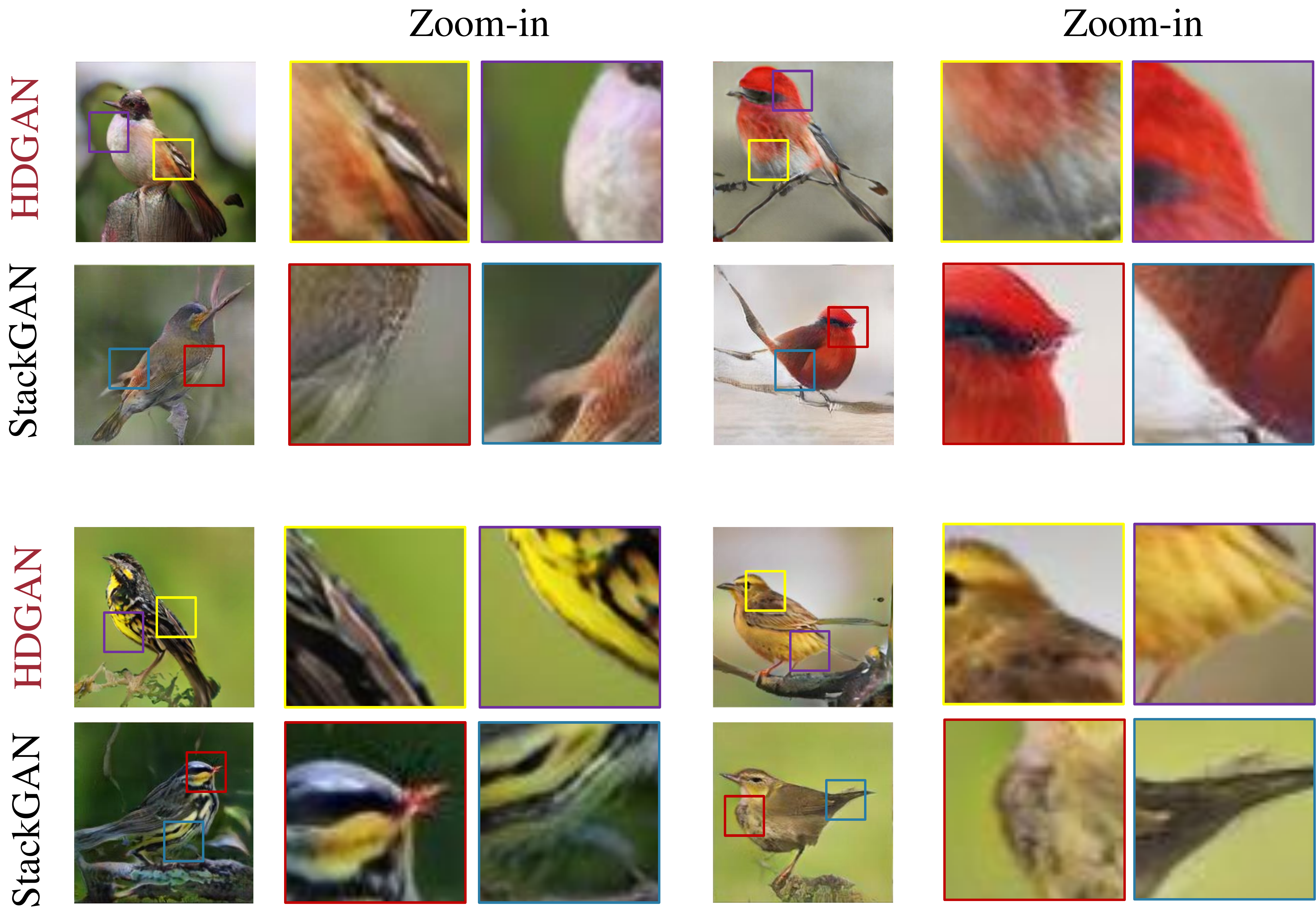}
	
	\caption{Zoomed-in samples compared with StackGAN. The best sample among 6 samples given an input text is selected. It can be clearly observed that our results show more smooth visual results. Especially, much less sharp pixel transitions exist in our results. }  
	\label{fig:bird3}
\end{figure*}
\begin{figure*}[ht!]
	\centering
	\includegraphics[width=0.99\textwidth]{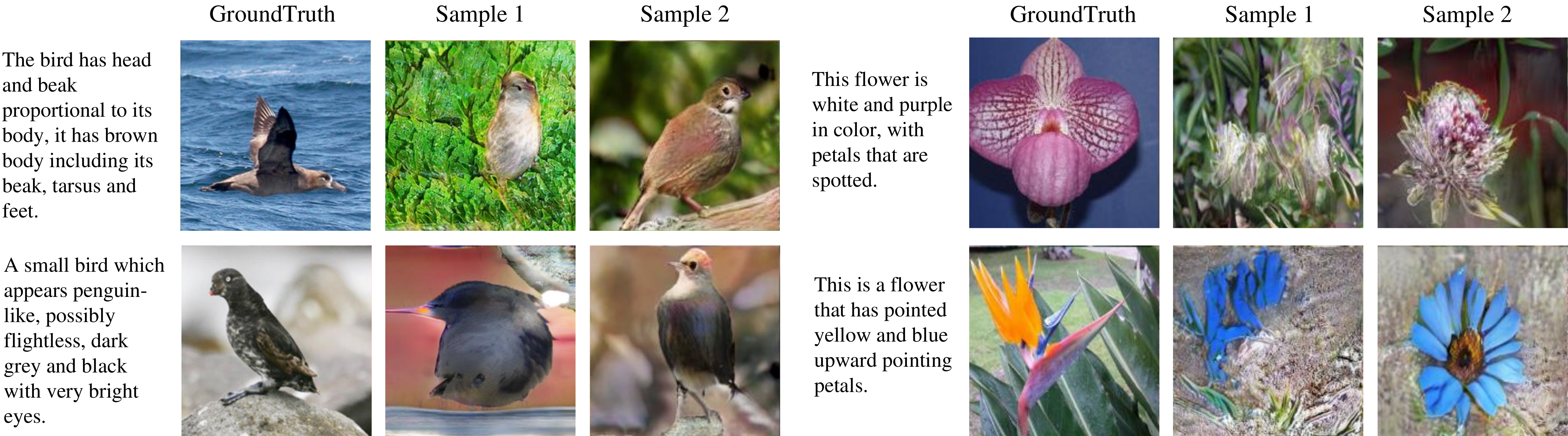}
	
	\caption{Illustration of failure cases due to artifacts, minor semantic inconsistency (compared with groundtruth), loss of object basic shapes. For each input text, we sampled ten samples. The figure shows a worst failed image (Sample 1) and a relative better image (Sample 2).}  
	\label{fig:fail}
\end{figure*}
\newpage
\begin{figure*}[th!]
	\centering
	\includegraphics[width=0.9\textwidth]{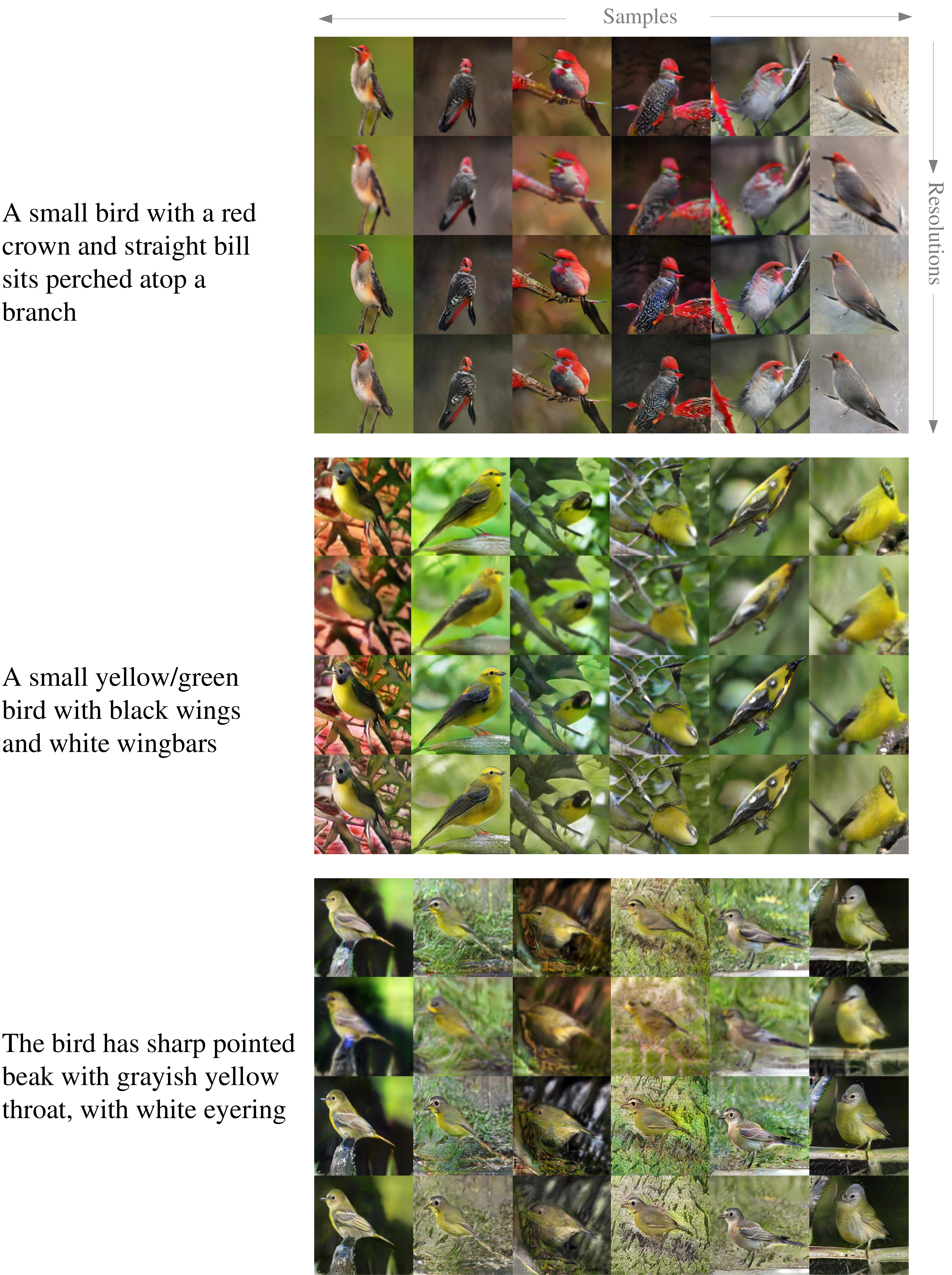}
	
	\caption{Sample results on CUB. For each input, 6 samples with resolutions of $64^2$, $128^2$, $256^2$, and $512^2$ are shown in 4 rows, respectively. }  
	\label{fig:bird}
\end{figure*}

\newpage


\begin{figure*}[th!]
	\centering
	\includegraphics[width=0.999\textwidth]{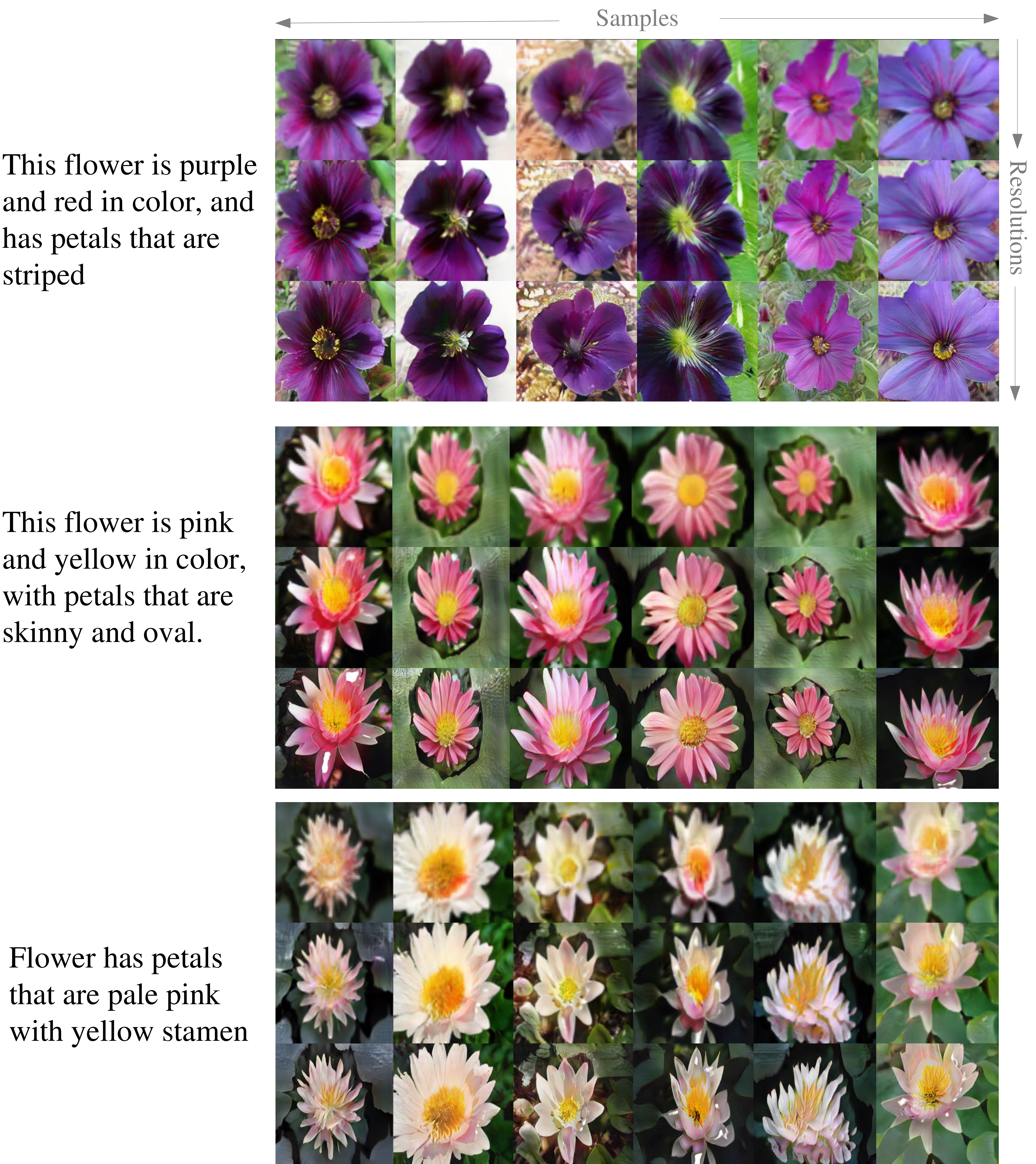}
	
	\caption{Sample results on Oxford-102. For each input,  6 samples with resolutions of $64^2$, $128^2$, and $256^2$ are shown in 3 rows, respectively. }  
	\label{fig:flower}
\end{figure*}
\newpage
\begin{figure*}[ht!]
	\centering
	\includegraphics[width=0.8\textwidth,width=0.8\textwidth]{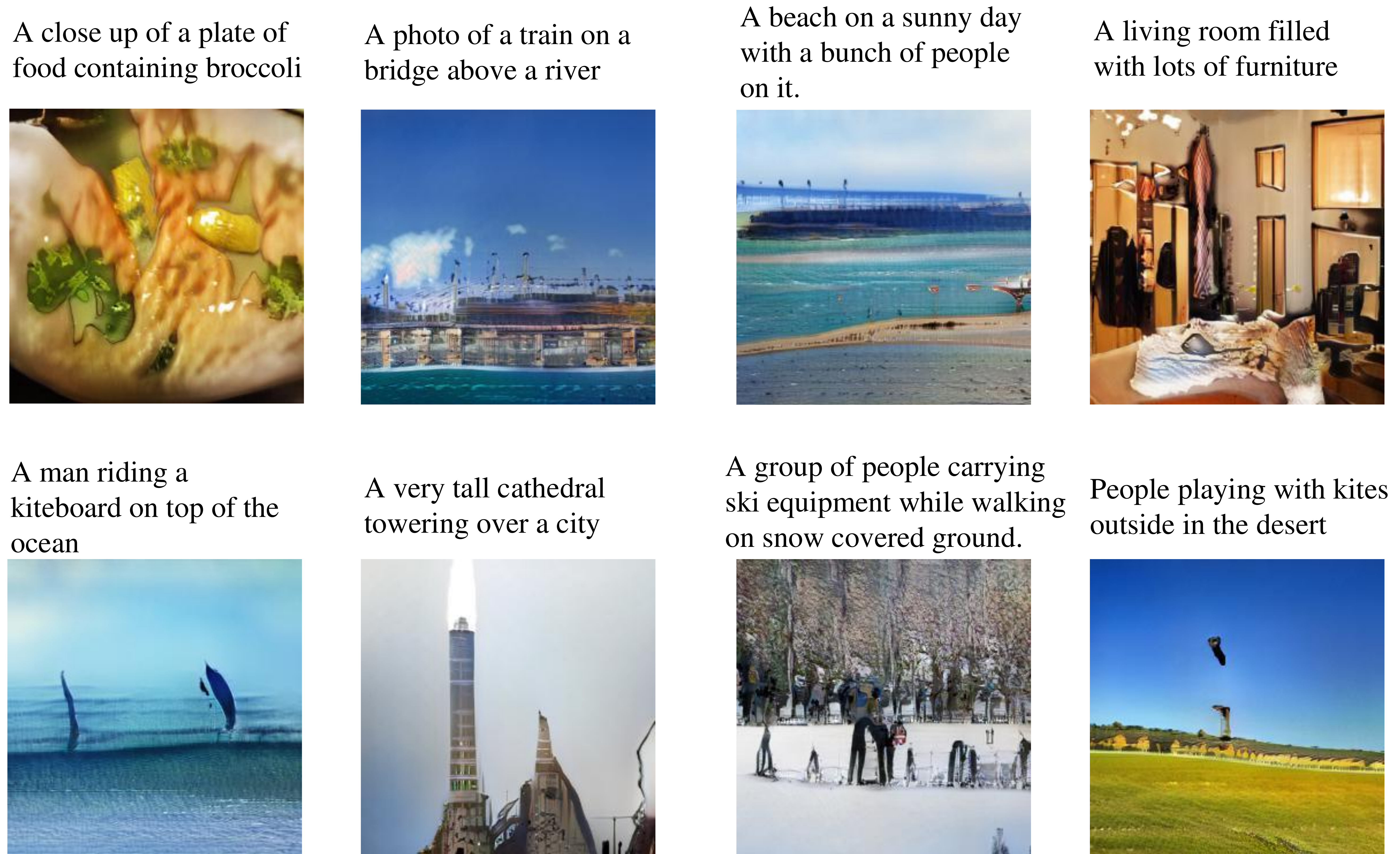}
	
	\caption{Sample results on COCO. We show 8 $256^2$ samples in very different scenes. }  
	\label{fig:coco}
\end{figure*}

\end{document}